\documentclass{article} %
\usepackage{etoolbox}       %

\usepackage{amsmath,amsthm}   %
\usepackage{mathtools}  %
\usepackage[dvipsnames]{xcolor}         %
\usepackage[utf8]{inputenc} 
\usepackage[T1]{fontenc}    

\newtoggle{arxiv}
\toggletrue{arxiv}
\iftoggle{arxiv}{
  \usepackage[parfill]{parskip}
  \usepackage[authoryear]{natbib}
  \setlength{\textwidth}{6.8in}  %
  \setlength{\textheight}{9in}
  \setlength{\oddsidemargin}{0in}
  \setlength{\evensidemargin}{0in}
  \setlength{\topmargin}{-0.5in}
  \newlength{\defbaselineskip}
  \setlength{\defbaselineskip}{\baselineskip}
  \setlength{\marginparwidth}{0.8in}
  \setlength{\parskip}{6pt}%
  \setlength{\parindent}{0pt}%

  \RequirePackage[T1]{fontenc}
  \RequirePackage[tt=false, type1=true]{libertine}
  \RequirePackage[varqu]{zi4}
  \RequirePackage[libertine]{newtxmath}

}{
  \usepackage{amsfonts}       %
  \usepackage{colm2024_conference}
  \addtolength{\tabcolsep}{-3pt} %
  \def\setstretch#1{\renewcommand{\baselinestretch}{#1}}
  \setstretch{0.98}
  \addtolength{\parskip}{-1pt}

  \usepackage[compact]{titlesec}
  \titlespacing{\section}{0pt}{*1}{*0}
  \titlespacing{\subsection}{0pt}{*1}{*0}
  \usepackage[subtle, mathdisplays=tight, charwidths=tight, leading=normal]{savetrees}
  \addtolength\textfloatsep{-0.5em}
  \addtolength\intextsep{-0.2em}
  \bibliographystyle{colm2024_conference}
}

\usepackage{booktabs}       
\usepackage{amsfonts}       
\usepackage{nicefrac}       
\usepackage{microtype}      
\usepackage{xcolor}         

\usepackage{multirow}
\usepackage{graphicx}
\usepackage{subcaption}
\usepackage{xcolor,colortbl}
\definecolor{Gray}{gray}{0.89}
\definecolor{LightCyan}{rgb}{0.88,1,1}
\usepackage{tabularx}
\usepackage{makecell}
\usepackage{wrapfig}
\usepackage{amsmath,amsthm}
\usepackage{mathtools, nccmath}
\usepackage{tablefootnote}
\DeclarePairedDelimiter{\nint}\lfloor\rceil
\DeclarePairedDelimiter{\abs}\lvert\rvert

\usepackage{threeparttable}

\usepackage{url}
\usepackage{hyperref}
\hypersetup{
     colorlinks   = true,
     linkcolor    = blue,
     citecolor    = magenta,
     urlcolor     = red,
     pdfborderstyle={/S/U/W 1},
}

\usepackage[capitalise,noabbrev]{cleveref}  %

\usepackage{pifont}%
\newcommand{\cmark}{\ding{51}}%
\newcommand{\xmark}{\ding{55}}%

\iftoggle{arxiv}{
  \title{Quamba2: A Robust and Scalable Post-training Quantization Framework for Selective State Space Models}
  \usepackage{authblk}
  \author[$^1$]{Hung-Yueh Chiang}
  \author[$^2$]{Chi-Chih Chang}
  \author[$^1$]{Natalia Frumkin}
  \author[$^3$]{\authorcr Kai-Chiang Wu}
  \author[$^2$]{Mohamed S. Abdelfattah}
  \author[$^1$]{Diana Marculescu}
  \affil[$^1$]{
    Chandra Family Department of Electrical and Computer Engineering, \protect\\ 
    The University of Texas at Austin
  }
  \affil[$^2$]{Department of Electrical and Computer Engineering, Cornell University}
  \affil[$^3$]{Department of Computer Science, National Yang Ming Chiao Tung University}
  \affil[ ]{{\texttt{\{hungyueh.chiang, nfrumkin, dianam\}@utexas.edu}}, \protect\\ {\texttt{\{cc2869, mohamed\}@cornell.edu}}, {\texttt{kcw@cs.nycu.edu.tw}}}
  \date{}
}{
  \title{Quamba2: A Robust and Scalable Post-training Quantization Framework for Selective State Space Models}
}

\begin{document}

\maketitle
\begin{abstract}
\noindent
State Space Models (SSMs) are emerging as a compelling alternative to Transformers because of their consistent memory usage and high performance. Despite this, scaling up SSMs on cloud services or limited-resource devices is challenging due to their storage requirements and computational power. To overcome this, quantizing SSMs with low bit-width data formats can reduce model size and benefit from hardware acceleration. As SSMs are prone to quantization-induced errors, recent efforts have focused on optimizing a particular model or bit-width for efficiency without sacrificing performance. However, distinct bit-width configurations are essential for different scenarios, like W4A8 for boosting large-batch decoding speed, and W4A16 for enhancing generation speed in short prompt applications for a single user.
To this end, we present Quamba2, compatible with \textbf{W8A8}, \textbf{W4A8}, and \textbf{W4A16} for both \textbf{Mamba1} and \textbf{Mamba2} backbones, addressing the growing demand for SSM deployment on various platforms. Based on the \emph{channel order preserving} and \emph{activation persistence} of SSMs, we propose an offline approach to quantize inputs of a linear recurrence in 8-bit by sorting and clustering for input $x$, combined with a per-state-group quantization for input-dependent parameters $B$ and $C$. To ensure compute-invariance in the SSM output, we rearrange weights offline according to the clustering sequence. The experiments show that Quamba2-8B outperforms two state-of-the-art SSM quantization methods and delivers 1.3$\times$ and 3$\times$ speed-ups in the pre-filling and generation stages, respectively, while offering 4$\times$ memory reduction with only a $1.6\%$ average accuracy drop. The evaluation on MMLU shows the generalizability and robustness of our framework. The code and quantized models will be released at: \url{https://github.com/enyac-group/Quamba}.
\end{abstract}

\section{Introduction}

State Space Models (SSMs) \citep{gu2020hippo, smith2023simplified, gu2023mamba, dao2024transformers}, offering \emph{constant} memory complexity, are emerging as efficient alternatives to Transformers \citep{vaswani2017attention} in various areas such as language modeling \citep{wang2024mambabyte, waleffe2024empirical}, vision \citep{zhu2024vision, liu2024vmamba, li2025videomamba}, and audio \citep{goel2022s, saon2023diagonal}. Some studies expand the size of the models and demonstrate their performance on par with Transformers of the same scale \citep{lieber2024jamba, team2024jamba, waleffe2024empirical}. However, the large size of SSMs limits the hardware options and increases the deployment costs.

Post-training quantization (PTQ) offers an attractive solution to efficient deployment by eliminating the needs of fine-tuning large models. PTQ reduces the bit-width of pre-trained weights and activations to lower-bit formats (such as 8-bit), cutting down memory use for weight storage and leveraging advanced hardware units. Recent studies \citep{xu2025mambaquant, chiang2024quamba} reveal that quantization techniques that are effective in Transformers struggle with SSMs due to the sensitivity of linear recurrence to quantization-induced errors. This prior work introduces PTQ algorithms tailored for SSMs to bridge the performance gap between low and half-precision models. However, they either do not explore diverse bit-widths \citep{chiang2024quamba} or fail to achieve satisfactory performance at lower bit-widths \citep{xu2025mambaquant}, such as W4A8.

Specific bit-width setups are crucial for certain scenarios. For instance, W4A8 enhances cloud service throughput with large-batch inputs \citep{lin2024qserve}, whereas W4A16 improves the efficiency of short prompt applications \citep{lin2024awq}. As a result, current SSM-based quantization methods \citep{xu2025mambaquant, chiang2024quamba} may underperform on edge devices or fail to maximize throughput on cloud services. Moreover, a recent study \citep{zhao2024qspec, kumar2025scaling, gong2024llmc} reveals that heavy quantization of model weights and activations (\emph{e.g.,} W4A4) impairs model generalization on multi-step reasoning tasks. Previous SSM-based studies overlook the generalizability of quantized models.

\begin{wrapfigure}{r}{0.55\textwidth}
\vspace{-30pt}
    \centering
    \includegraphics[width=.53\textwidth]{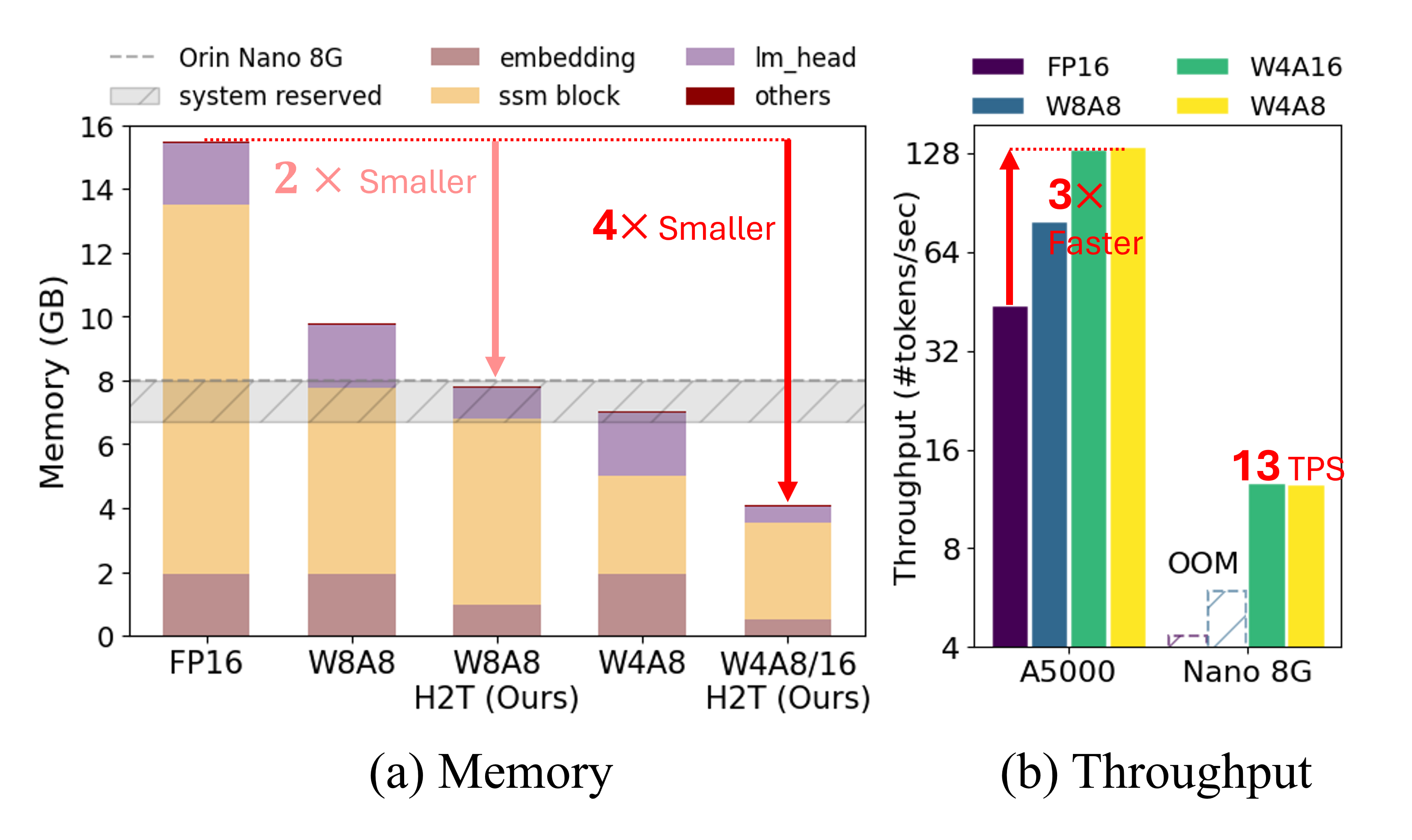}
    \caption{\textbf{(Quamba2-8B memory and throughput.)} The head-to-toe (H2T) quantization enables the deployment of Mamba2-8B on edge platforms. Quamba2 delivers $3\times$ throughput on Nvidia A5000 and 13 tokens-per-second (TPS) on Nvidia Nano 8G.}
    \label{fig:mamba2_8b_memory}
\end{wrapfigure}

\begin{figure*}[t!]
    \centering
    \includegraphics[width=\textwidth]{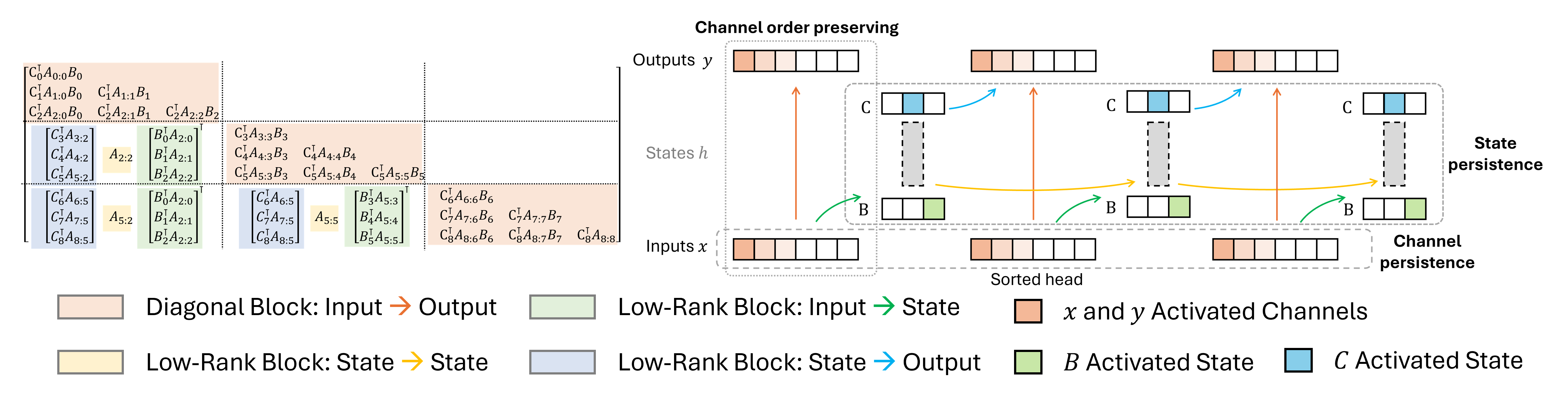} 
    \caption{\textbf{(SSD flows with sorted heads and the activation persistence.)} We sort the head channels prior to applying quantization scaling factors. The orange blocks on the right indicate the activated channels with higher values in the input and output SSD heads. The SSD performs \emph{channel-wise} calculation thereby retaining the channel order between input $x$ and output $y$, which we call \emph{channel order preserving}. The blue and green blocks represent the activated states of input-dependent parameters $B$ and $C$. Our study shows that activated channels and states remain \emph{consistent} across time steps and input samples, a property we denote as \emph{channel persistence} and \emph{state persistence}.}
    \label{fig:overview}
\end{figure*}

\begin{figure*}[t!]
    \centering
    \includegraphics[width=\textwidth]{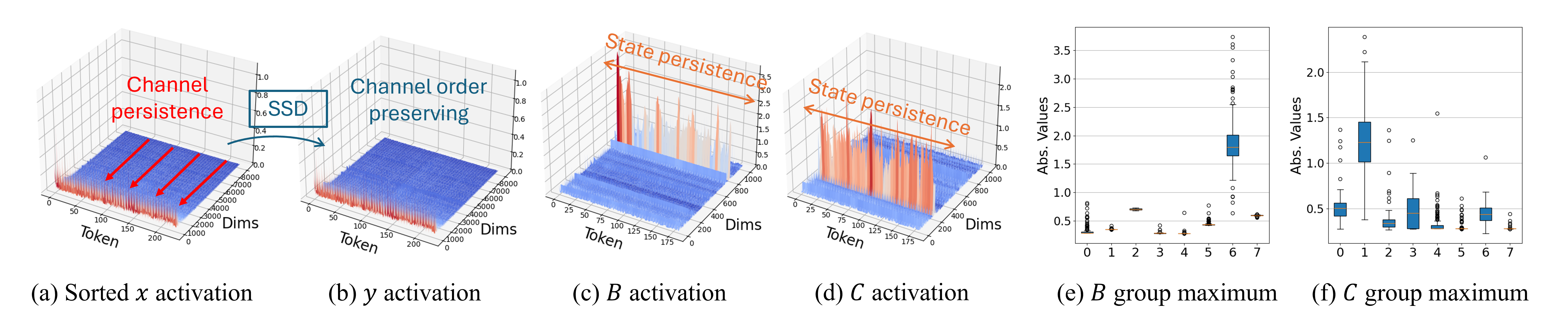} 
    \caption{\textbf{(Channel order preserving and activation persistence.)}  We show the activations in the last block of Mamba2-8B. For an input with $t$ tokens, we demonstrate that the $x$ remains sorted by the maximum of the calibrated channel (a). The SSD calculation is \emph{channel-wise}, so the output channel order $y$ matches the input order $x$ (b). For $B$ and $C$, the activated states remain consistent over time steps $t$ (c-d) and input samples (e-f). We leverage the observations and design our techniques, \emph{sort-and-cluster} and \emph{per-state-group quantization}, to increase the quantization precisions for $x$ (a), $B$, and $C$ (c-f). }
    \label{fig:persistent}
\end{figure*}

To address these issues, we present Quamba2, a robust and scalable post-training quantization framework for selective SSMs. As shown in Table \ref{tab:supported_models} and \ref{tab:supported_bitwidth}, our framework supports \emph{head-to-toe} \textbf{W8A8}, \textbf{W4A8}, and \textbf{W4A16} for both \textbf{Mamba1} \citep{gu2023mamba} and \textbf{Mamba2} \citep{dao2024transformers}, meeting the demands for SSM deployment on cloud and edge platforms. Based on \emph{channel order preserving} and \emph{activation persistence} of the SSM computation, as shown in Figure \ref{fig:overview} and \ref{fig:persistent}, we employ an offline \emph{cluster-aware} weight reordering approach to group SSM heads and channels with similar value ranges, allowing them to share a quantization scaling factor and boost quantization precision. For selective SSM input-dependent parameters ($B$, $C$), we identify \emph{state persistence} in activations and apply quantization per state group.
Our sort-and-cluster and per-state-group quantization methods improve quantization accuracy, closing the accuracy gap in half-precision models. In Figure \ref{fig:mamba2_8b_memory} and the rest of our experiments, we show that Quamba2-8B surpasses two leading SSM quantization methods, achieving up to 1.3$\times$ and 3$\times$ higher speeds in prefilling and generation, respectively, and offering a 4$\times$ memory reduction, with only a $1.6\%$ accuracy loss across six zero-shot tasks. Additionally, we tested Quamba2 on MMLU \citep{hendrycks2020measuring}, a large multitasking dataset, demonstrating the generalizability and robustness of our framework.

\begin{table*}[t]
\centering
\setlength{\tabcolsep}{1.5pt}
\begin{minipage}{0.48\textwidth}
    \centering
    \caption{\textbf{(Supported models.)} Our framework supports W8A8, W4A8, and W4A16 for both Mamba1 \citep{gu2023mamba} and Mamba2 \citep{dao2024transformers}.}
    \resizebox{\textwidth}{!}{
    \begin{tabular}{@{}l|cc|ccc@{}}
    \toprule
    \multirow{2}{*}{Methods}                      & \multicolumn{2}{c|}{Models}       &  \multicolumn{3}{c}{Bitwidth}     \\
                                                  & Mamba1 & Mamba2 &  W8A8  & W4A8   & W4A16      \\ \midrule\midrule
    MambaQuant (\citeauthor{xu2025mambaquant})    & \cmark &     -  & \cmark & \cmark & -          \\ \midrule
    Quamba (\citeauthor{chiang2024quamba})        & \cmark &     -  & \cmark & -      & -          \\ \midrule
    Quamba2 (Ours)                                & \cmark & \cmark & \cmark & \cmark & \cmark \\ \bottomrule  
    \end{tabular}
    }
    \label{tab:supported_models}
\end{minipage}
\hfill
\begin{minipage}{0.48\textwidth}
    \centering
    \caption{\textbf{(Supported bit-widths.)} Quamba2 supports head-to-toe (H2T) 4/8-bit from the embedding layer, SSM blocks, to the final output layer (\emph{i.e.,} lm head).}
    \resizebox{\textwidth}{!}{
    \begin{tabular}{@{}l|c|c|c|c@{}}
    \toprule
    Methods                                     & Embed.   & SSM blocks & lm head  & H2T     \\ \midrule\midrule
    MambaQuant (\citeauthor{xu2025mambaquant})  &  16-bit  & 4/8-bit    &  16-bit  & \xmark  \\ \midrule
    Quamba (\citeauthor{chiang2024quamba})      &  16-bit  & 8-bit      &  16-bit  & \xmark  \\ \midrule
    Quamba2 (Ours)                              &  4/8-bit & 4/8-bit    &  4/8-bit & \cmark  \\ \bottomrule
    \end{tabular}
    }
    \label{tab:supported_bitwidth}
\end{minipage}
\end{table*}

\section{Related Work}

\paragraph{Model quantization.}
Representing model weights and activations with low bit-width data types reduces the cost of storing and loading parameters and benefits from advanced low bit-width computing units (\emph{i.e.,} Tensor Cores). Quantization methods are generally divided into two categories: Quantization-aware training (QAT) \citep{liu2023llm, dettmers2024qlora, yu2025slender, tang2024bi} and post-training quantization (PTQ) \citep{zhu2024survey, zhou2024survey}. QAT requires additional GPU resources and training efforts to adapt models to low bit-width. PTQ is an attractive option for large language models (LLMs) since it eliminates the need for training. Our work falls under PTQ and minimizes GPU requirements. Our framework provides bit-width configurations of W8A8, W4A8, and W4A16 for SSM-based language models, delivering generic memory and latency reduction on all target platforms.

\vspace{-5pt}
\paragraph{PTQ and weight reordering for Transformers.}
Post-training quantization (PTQ) techniques are generally classified into two categories: weight-only quantization (\emph{e.g.,} W4A16) and weight-activation quantization (\emph{e.g.,} W8A8) \citep{zhu2024survey}. Weight-only quantization \citep{frantar2023gptq, lin2024awq} minimizes weight storage, while weight-activation quantization \citep{zhao2024atom, ashkboos2024quarot} optimizes throughput with low bit-width operations. Reordering weights is frequently used to enhance quantization precision \citep{zhao2024atom, yuan2023rptq} or efficiency \citep{lin2024qserve} of Transformers, but its use and its subsequent effectiveness in SSMs is unclear. Our study shows that the selective State Space Duality (SSD) computing \citep{dao2024transformers} \emph{preserves channel order} between input and output, with activated channels and states \emph{consistent} over time.

\vspace{-5pt}
\paragraph{PTQ and mixed-precision for SSMs.}
\citet{xu2025mambaquant} and \citet{chiang2024quamba} highlight that standard quantization techniques for Transformers are not effective for SSMs and propose PTQ algorithms tailored for SSMs. 
Despite this, these strategies do not offer a variety of bit-width configurations \citep{chiang2024quamba} and struggle to perform well at reduced bit-widths such as W4A8 \citep{xu2025mambaquant}.
Moreover, \citet{zhao2024qspec} show that 4-bit models lose generalizability, and \citet{kumar2025scaling} indicate the best performance under memory constraints for a bit-width of 6-8, with worse results for a bit-width of 4.
Also, previous mixed-precision research focuses soly on Convolutional Neural Networks (CNNs) \citep{wang2019haq, dong2019hawq} and Transformers \citep{zhao2021automatic}.
We aim to fill the missing point of low bit-width and mixed-precision SSMs.
Our framework provides \textbf{W8A8}, \textbf{W4A8}, and \textbf{W4A16} for both \textbf{Mamba1} \citep{gu2023mamba} and \textbf{Mamba2} \citep{dao2024transformers} with practical speed-up and memory reduction, addressing the growing demand for the deployment of SSMs both in the cloud and on the edge.
We evaluate Quamba2 on a large and challenging multitasking dataset, MMLU \citep{hendrycks2020measuring}, to show the robustness of our framework.

\section{Background}

\subsection{Model Quantization}

\paragraph{Notations.}
We follow the notation in \citet{chiang2024quamba}.  
We use $X$ to represent the floating-point matrices, and $\overline{X}$ to represent their quantized matrices with their floating-point scaling factors $s_x$.
For operators, we use $\overline{f}(\cdot)$ to represent the quantized version of the function $f(\cdot)$ (\emph{i.e.,} the weights are quantized in the function $\overline{f}$).

\paragraph{Quantization.}
We focus on \emph{symmetric uniform quantization} to approximate floating-point weights and activations with discrete $N$-bit signed integers (\emph{i.e.,} INT8 or INT4) due to its \emph{hardware compatibility}.
The general symmetric uniform quantization function is defined as 
\begin{equation} \label{eq:quantization}
\overline{X} = \text{\footnotesize{Clamp}}\Big(\nint*{\frac{X}{s}}, -2^{N-1}, 2^{N-1}-1\Big),
\end{equation}
where $s = \text{\footnotesize{Max}}\big(\abs*{X}\big) / (2^{N-1}-1)$. $\overline{X}$ represents the quantized weights or activations, $X$ is the input matrix in floating point, and $s$ is the scaling factor (\emph{i.e.,} quantization step) that is determined by the target bit-width $N$ ($N=\{4, 8\}$ in our setting).
The \emph{static} scaling factor $s$ is pre-calibrated and \emph{fixed} during inference.

\begin{figure*}[t!]
    \centering
    \includegraphics[width=\textwidth]{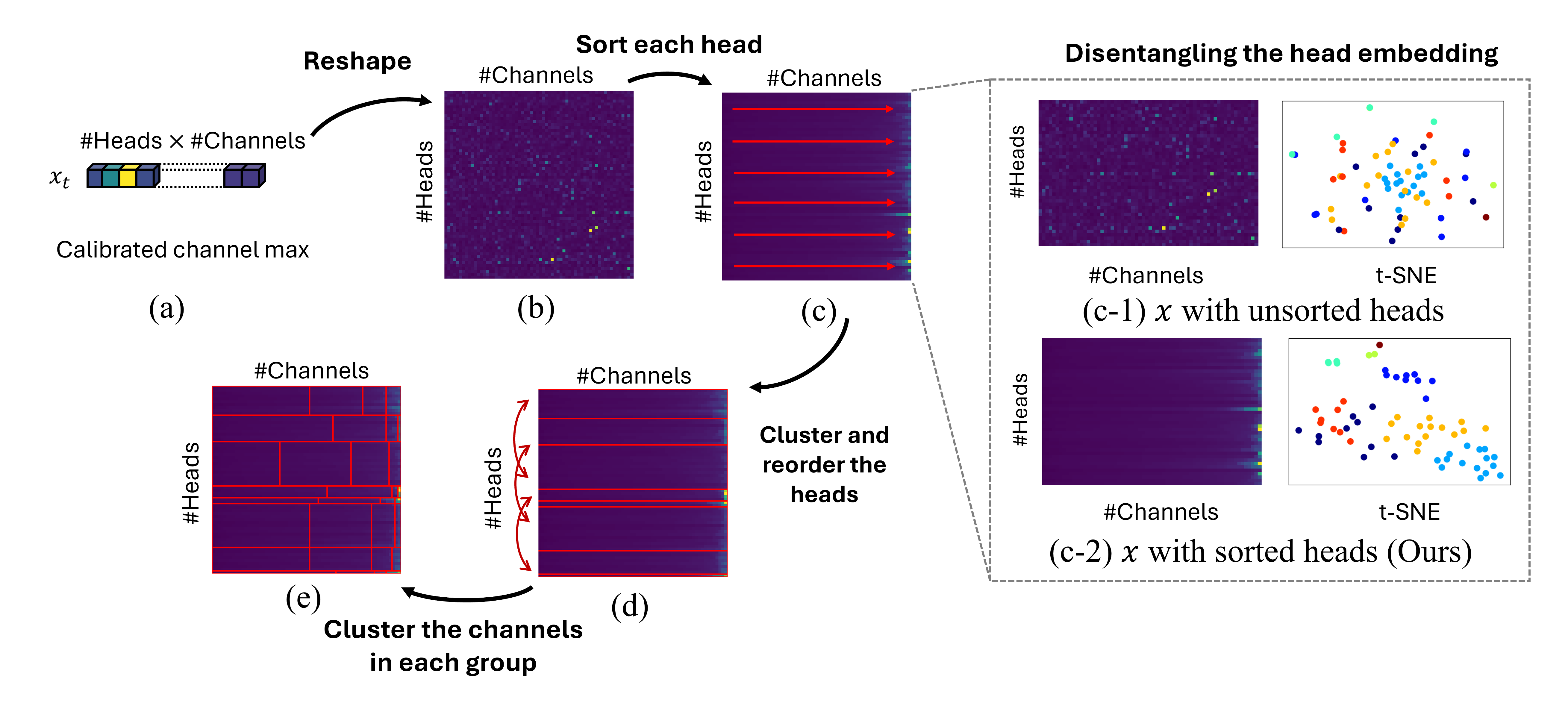} 
    \caption{\textbf{(Sort-and-cluster.)} We leverage the \emph{channel-persistent} property in SSMs to sort the channel with the calibrated maximum (a-c). The sorted heads disentangle the embedding, as shown in (c-1) and (c-2), enabling the clustering on the heads.
    We cluster the sorted heads into $m$ groups ($m=8$ in (d)), and reorder the weights offline to match the clustering results. Then, we apply the clustering again in each head group to cluster the channels into $n$ groups ($n=4$ in (e)). For each group, a scaling factor is calculated, resulting in $m \times n$ factors used to quantize $x_t$ to 8-bit.}
    \label{fig:clustering}
\end{figure*}

\subsection{Selective State Space Models}
\begin{wrapfigure}{r}{0.5\textwidth}
\vspace{-0pt}
    \centering
    \includegraphics[width=.48\textwidth]{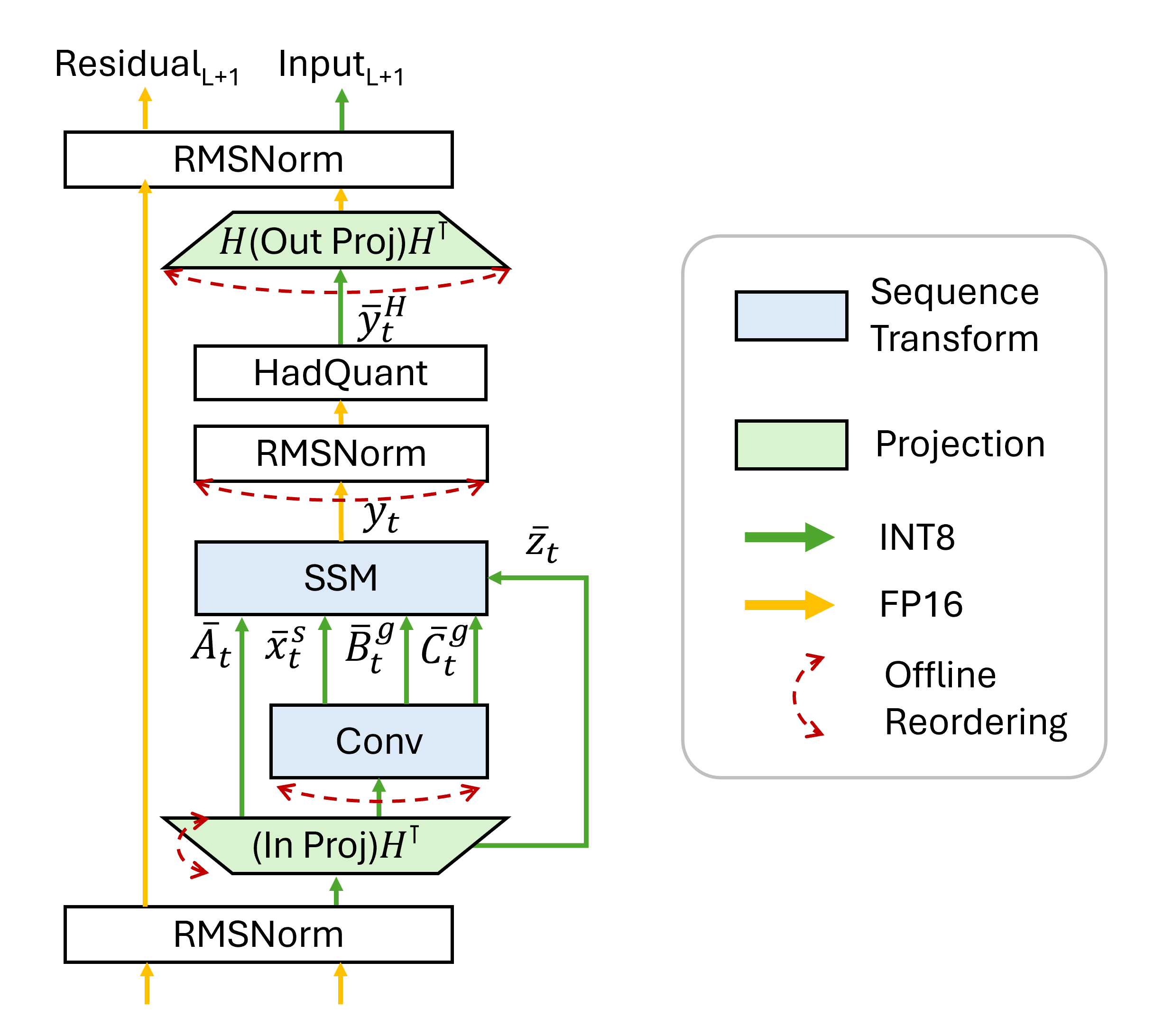}
    \caption{\textbf{(Quamba2 precision.)} The detailed precision mapping of W4A8 and W8A8 Quamba2. We reorder the weights \emph{offline} to match the sorting and clustering indices of $\bar{x}^s_t$, and apply per-state-group quantization on $\bar{B}^g_t$ and $\bar{C}^g_t$.}
    \label{fig:mamba2_precision}
\end{wrapfigure}

The selective SSM \citep{gu2023mamba, dao2024transformers} transforms the time-invariant SSM \citep{gu2020hippo} to a time-varying system.
The system dynamics is defined by
\begin{equation} \label{eq:ssm}
\begin{aligned}
h_t = \dot{A_t} h_{t-1} + \dot{B_t} x_t, \quad
y_t = C_t h_t + D x_t
\end{aligned}
\end{equation}
where $(\dot{A_t}, \dot{B_t}, C_t)$ are input-dependent.
$\dot{A_t}$ and $\dot{B_t}$ are discrete parameters of $A$ and $B$.
The discretization function for $\dot{A_t}$ and $\dot{B_t}$ with a given input-dependent $\Delta_t$ is defined as $\dot{A_t} = \exp(\Delta_t A)$, $\dot{B_t} = (\Delta_t A)^{-1} (\exp(\Delta_t A) - I) \cdot \Delta_t B_t \approx \Delta_t B_t$.
$(A, D)$ are trainable parameters, and $D$ is an optional residual parameter.
An optional residual branch $z_t$ is applied to the SSM output such that $ y_t\cdot\text{SiLU}(z_t) $ before the output projection.
We follow \citet{dao2024transformers} and abstract the selective SSM computation at the time step $t$ with the function
\begin{equation} \label{eq:ssm_function}
y_t = \text{SSM}(\dot{A_t}, \dot{B_t}, C_t)(x_t) .
\end{equation}
Optional $z_t$ and $D$ are omitted in the function.
We omit the subscript $t$ to represent the computation for the entire sequence.
The abstract SSM block is shown in Figure \ref{fig:mamba2_precision}.

\paragraph{Mamba1.} 
\citet{gu2023mamba} presents selective SSMs in which the parameters $B$, $C$, and $\Delta$ vary with input (\emph{i.e.,} time-varying), allowing the model to selectively prioritize or ignore inputs based on their content. The interaction with the input $x_t$ is specified as $B_t = \text{F}_B(x_t), \quad C_t = \text{F}_C(x_t), \quad \Delta_t = \text{softplus}(\text{F}_\Delta(x_t))$, where $\text{F}_B$ and $\text{F}_C$ are linear transformations mapping $x_t$ to $B_t$ and $C_t$. The function $\text{F}_\Delta$ involves two sequential projection layers, formulated as $\text{F}_\Delta = \text{Proj}(\text{Proj}(x_t)) + \text{bias}$. The $x_t$ is calculated from the input of the block $u_t$ with a projection layer at the time step $t$.

\paragraph{Mamba2.}
\citet{dao2024transformers} establish a theoretical link, Structured State Space Duality (SSD), between selective SSMs and self-attention.
They also introduce an efficient algorithm that utilizes matrix multiplication units on contemporary hardware to perform linear recurrence calculations.
Mamba2 simplifies block design by removing sequential linears where $x_t, B_t, C_t$, and $\Delta_t$ are produced in parallel with a single projection layer such that $(x_t, B_t, C_t, \Delta_t) = \text{F}(u_t)$, where $u_t$ is the block input at the time step $t$.
The modified block design is better suited to tensor parallelism \citep{shoeybi2019megatron} in the context of larger models.


\subsection{Quantizing Selective SSMs}

\paragraph{SSM input parameters.}
The SSM defined in Equation \ref{eq:ssm_function} receives an input in the form of $(\dot{A_t}, \dot{B_t}, C_t; x_t)$. Recent efforts \citep{xu2025mambaquant, chiang2024quamba} show that the SSM block is extremely sensitive to quantization-induced errors in $x_t$ due to the linear recurrence mechanism in Mamba1 \citep{gu2023mamba}. \emph{Our work indicates that the phenomenon persists in Mamba2} \citep{dao2024transformers}. To address this issue, we propose \emph{sort-and-cluster} to quantize the input $x_t$ with 8-bit. Our method groups the channels across the heads with the same value range to create a smoother landscape in the group, and therefore increases the quantization precision.

\paragraph{SSM outliers.}
Prior studies on Transformers \citep{dettmers2022gpt3, xiao2023smoothquant} have detected channel-persistent outliers. A common method for outlier elimination is applying the Hadamard transform \citep{ashkboos2024quarot, liu2024spinquant}. In SSM quantization \citep{xu2025mambaquant, chiang2024quamba}, \emph{online} Hadamard matrices transform the input of output projection into a smoother space, enhancing the quantization precision. Although the fast Walsh–Hadamard transform (FWHT) can be executed in parallel with a $n\text{log}n$ complexity \citep{tridao_had, sloane1999library}, we adhere to \citet{xu2025mambaquant} and \citet{chiang2024quamba} to quantize the output projection input, with the aim of minimizing \emph{online} Hadamard transform overheads.

\section{Proposed Method: Quamba2}

\subsection{Quantizing SSM Parameters}
Our method is based on two findings in SSM activations: \emph{channel persistence} and \emph{state persistence}, together with a computational property of SSM: \emph{channel order preserving}.
The notation follows the definition from Equation \ref{eq:ssm_function}.


\vspace{-3pt}
\paragraph{Sort-and-cluster.}
We observe the \emph{persistence} of the channel magnitude and the \emph{preservation} of channel order in the SSM input $x$ and output $y$, as shown in Figure \ref{fig:overview}.
Although $x$ is sensitive to quantization-induced errors in Mamba2 \citep{dao2024transformers}, with the findings of \citet{chiang2024quamba} still applicable, \citet{chiang2024quamba} overlook the persistence characteristic and order-preserving of the SSM channel.
In contrast, we leverage these two properties to first sort the head channels and group both heads and channels.
Specifically, we first obtain the channel maximum from a calibration dataset.
In Figure \ref{fig:persistent} (a), we visualize the $x$ sorted by the \emph{offline} calibrated channel maximum of the last block of Mamba2-8B.
$x$ remains sorted input with an \emph{online} $t$-token sample.
The sorted $x$ disentangles the head embedding, allowing head grouping.
Figure \ref{fig:clustering} (c1-c2) shows that heads with similar characteristics are closely grouped, leading to the use of unsupervised clustering into $m$ groups.
For each group of heads, we apply the clustering algorithm again to group channels into $n$ groups.
The scaling factor is calculated for every group, leading to a total of $m \times n$ scaling factors, which are then utilized to quantize $x_t$ to 8-bit precision.
The detailed \emph{sort-and-cluster} process is shown in Figure \ref{fig:clustering}.
We find that $m=4$ and $n=4$ provide sufficiently good results throughout all experiments.
The $\bar{x}^s_t$ in Figure \ref{fig:mamba2_precision} refers to the activation applied with sort-and-cluster.

\vspace{-3pt}
\paragraph{Per-state-group quantization.}
\citet{dao2024transformers} relax the number of state group size and introduce a Multi-input SSM where $B_t$, $C_t$ matrices are shared across all channels of the input $x_t$, akin to grouped-query attention \citep{ainslie2023gqa} in Transformers.
Our findings indicate that the activated states (with larger numerical values) are the same across time steps $t$ and input samples.
In Figure \ref{fig:persistent} (c-f), we visualize the activation distribution of $B$ and $C$ in the last block of Mamba2-8B.
The number of groups in $B$ and $C$ is set to $8$, where each group has $128$ channels.
Figure \ref{fig:persistent} (c-d) shows that only a few groups are activated with larger values.
For example, in Figure \ref{fig:persistent} (e-f), group six in $B$ is mostly activated, while group seven in both $B$ and $C$ has minimal variations.
Thus, we apply per-state-group quantization to $B$ and $C$, where each group utilizes a single scaling factor.
The $\bar{B}^g_t$ and $\bar{C}^g_t$ in Figure \ref{fig:mamba2_precision} refer to the activations applied with per-state-group quantization.
The per-state-group quantization largely increases the quantization precision in the groups where the value range is small, \emph{e.g,} group seven in both $B$ and $C$.
We show that per-state-group quantization is key to mitigating the performance gaps with the FP16 model for Mamba2-8B.

\subsection{System and Framework Design}
\paragraph{Cluster-aware weight reordering.}
We create a new channel and head sequence in \emph{sort-and-cluster}, where the heads within the same cluster are grouped and their channels are arranged by the pre-calibrated maximum.
To produce the activations with the sorting and clustering orders, we use clustering and sorting indices to reorder \emph{offline} the input projection, causal convolution, normalization, and output projection in the block.
The output column of input projection weights and the channel of causal convolution weights are reordered.
As SSD computing maintains channel order (see Figure \ref{fig:overview} right), we reorder normalization weights and apply fused Hadamard quantization. 
Finally, input rows of the output projection are rearranged using the same indices to keep the output the same.
The \emph{offline} cluster-aware weight reordering is depicted in Figure \ref{fig:mamba2_precision}.

\begin{table}[t]
\centering
\small
\begin{minipage}{0.48\textwidth}
    \centering
    \caption{\textbf{(SSD latency.)} We profile SSD latency of Mamba2-8B in milliseconds (\emph{ms}) across sequence lengths with different input bit-width. We set batch size to eight.}
    \setlength{\tabcolsep}{7pt}
    \resizebox{\textwidth}{!}{ 
        \begin{tabular}{lcccc}
            \toprule
            Inputs & $L=256$ & $512$ & $1024$ & $2048$ \\
            \midrule\midrule
            FP16        & 0.82        & 1.61        & 3.51        & 7.22          \\
            Int8 (Ours) & 0.76        & 1.47        & 2.97        & 6.07          \\ \midrule
            Speedup     & $1.08 \times$ & $1.10 \times$ & $1.18 \times$ & $1.19 \times$   \\
            \bottomrule
        \end{tabular}
    }
    \label{tab:ssd_latency}
\end{minipage}
\hfill
\begin{minipage}{0.48\textwidth}
    \centering
    \caption{\textbf{(Quamba2 model size in GB.) } We profile the model size in GB of different bit-width configurations for Mamba1 and Mamba2 in our framework.}
    \resizebox{\textwidth}{!}{
    \begin{tabular}{@{}l|c|c|ccc@{}}
        \toprule
        Models                  & Size & FP16 & W8A8 & W4A8 & W4A16 \\ \midrule\midrule
        Mamba1                  & 2.8B & 5.3 GB  & 2.8 GB & 1.5 GB & 1.5 GB  \\ \midrule
        \multirow{2}{*}{Mamba2} & 2.7B & 5.2 GB  & 2.7 GB & 1.4 GB & 1.4 GB  \\ \cmidrule{2-6}
                                & 8B   & 15.7 GB & 7.9 GB & 4.0 GB & 4.0 GB  \\ \bottomrule
    \end{tabular}
    }
    \label{tab:model_size}
\end{minipage}
\end{table}

\paragraph{Offline Hadamard matrix fusion.}
Hadamard matrices have the computational property $\mathbf{H}_n\mathbf{H}_n^\top = n \mathbf{I}_n$ where $n$ denotes $n$-dimensional square matrices.
We therefore fuse \emph{offline} the Hadamard matrices into the input and output linear projections.
For the output projection, the Hadamard matrices are multiplied at both sides of the weight matrix, such that $\mathbf{W}_{\text{out}}^H = \mathbf{H}_n \mathbf{W}_{\text{out}}  \mathbf{H}_n^\top$.
We fuse a Hadamard matrix at the input side of the input projection weight, such that $\mathbf{W}_{\text{in}}^H = \mathbf{W}_{\text{in}}  \mathbf{H}_n^\top$.
Thus, pairing Hadamard matrices in input/output projections with online Hadamard quantization results in compute-invariance \citep{ashkboos2024slicegpt, ashkboos2024quarot}, yielding an identical block output.
The \emph{offline} Hadamard matrix fusion is shown in Figure \ref{fig:mamba2_precision}.
We apply the 4-bit/8-bit quantization on the weights after matrix fusion.

\vspace{-3pt}
\paragraph{Efficient 4-bit/8-bit Mamba blocks.}
Our framework accommodates W8A8, W4A8, and W4A16 projection kernels, a W8A8 causal convolution kernel, 4-bit and 8-bit embedding kernels, and 8-bit selective scan and SSD kernels.
For projection layers, we reorder the weights and their per-group scaling factors \citep{lin2024qserve, frantar2024marlin, zhang2024qqq} to maximize the Tensor Core loading throughput.
The output scaling factors are fused to the input scaling factors such that $\bar{Y}=s_W s_{\text{fused}}\bar{W}\bar{X}$ where $s_{\text{fused}} = s_X/s_Y$.
We implement W4A8 and W4A16 \emph{fused matmul-transpose} kernels for the Mamba1 block.
For sequence transformations, we load the 8-bit activations and 8-bit cached states to reduce memory pressure, thus improving latency, as shown in Table \ref{tab:ssd_latency}.
In the forward Hadamard transform, the scaling factor $s_y$ is integrated, making $\overline{y}^H = \frac{1}{s_y}\mathbf{H}_ny$, thereby avoiding extra computational load during quantization.
The efficient kernels of our framework provide \emph{generic speed-up} and \emph{memory reduction}, addressing the increasing demands for the deployment of SSM on the cloud and on the edge.

\vspace{-3pt}
\paragraph{Head-to-toe quantization.}
Quantizing from embedding to the output head (\emph{i.e.,} Head-to-toe quantization) brings additional memory and latency reduction, which is necessary on edge computing platforms with limited memory capacity.
As shown in Figure \ref{fig:mamba2_8b_memory}, our head-to-toe (H2T) quantization enables the deployment of Mamba2-8B on Nano 8G.
Specifically, we employ \emph{per-token} quantization to the embedding layer, and \emph{per-group} quantization to the weight of the head.
As shown in Table  \ref{tab:supported_bitwidth}, we implement the CUDA kernels and support the 4-bit/8-bit embedding layer and 4-bit/8-bit output head.
Therefore, our framework achieves generic 4$\times$ memory reduction.

\vspace{-3pt}
\paragraph{Improving robustness via W4A$X$-mixed.}
\citet{zhao2024qspec} demonstrate that applying W4A4 to all blocks compromises generalizability of Transformers. We extend such analysis to verify SSM robustness and generalizability on MMLU \citep{hendrycks2020measuring} dataset.  Our findings indicate that while full W4A8 quantization maximizes prefilling speedup, it suffers from a notable generalization gap ($-5.8\%$ on MMLU \emph{vs.} $-2.1\%$ on LAMBADA). In contrast, full W4A16 quantization demonstrate robustness but comes at the cost of increased prefilling latency.
To address this, we introduce mixed-precision support in our framework. We automatically search salient blocks based on their performance sensitivity  and assign them a higher precision. Our W4A\{8/16\}-mixed SSM achieves a 2.9\% accuracy improvement on MMLU while incurring only a 10\% increase in prefilling latency.

\vspace{-5pt}
\section{Experiments}

\subsection{Experimental Setup}
We provide framework design details in Appendix \ref{sec: Details of Quamba2 Framework}.

\vspace{-3pt}
\paragraph{Evaluations.}
We use LM-EVAL \citep{eval-harness} to evaluate Quamba2 and baselines on six zero-shot downstream tasks: LAMBADA \citep{LambdataDataset}, HellaSwag \citep{HellaSwag}, PIQA \citep{PIQADataset}, ARC \citep{arcDataset} and WinoGrande \citep{WinoGrandeDataset}, and show the average accuracy over five runs in each table.
To compare with MambaQuant \citep{xu2025mambaquant}, we average the accuracy across five datasets: ARC-easy, ARC-challenge, PIQA, WinoGrande and HellaSwag.
The full evaluation is in Appendix Section \ref{Appendix:full_zero_shot_evaluation}, where we follow the evaluation protocol in Mamba1 \citep{gu2023mamba}, and report the accuracy for LAMBADA, WinoGrande, PIQA, and ARC-easy, and accuracy normalized by sequence length for HellaSwag and ARC-challenge.
To show generalizability and robustness, we evaluate the 8B models on MMLU \citep{hendrycks2020measuring}, a large multitask test consisting of multiple-choice questions from various domains.

\vspace{-3pt}
\paragraph{Baselines.} 
In our W8A8 setting, we compare our framework with the latest quantization methods for SSM, MambaQuant \citep{xu2025mambaquant} (W8A8, W4A8) and Quamba \citep{chiang2024quamba} (W8A8) on zero-shot downstream tasks.
In the Quamba setting \citep{chiang2024quamba}, we applied the Hadamard transform to the output projection input and implemented percentile clipping on the input SSM, establishing our W8A8 Mamba2 baseline for latency and accuracy.
We also provide the latency for W4A8 and W4A16.

\begin{wraptable}{r}{0.48\textwidth} 
\centering
\caption{\textbf{(Mamba2-8B latency.)} TPOT and TTFT on Nvidia A5000 GPU and Orin Nano 8G are measured in milliseconds (\emph{ms}) with one batch. TTFT is profiled with 1024 tokens. (OOM: out-of-memory)}
\label{tab:quamba_latency}
\renewcommand{\arraystretch}{1.3} 
\begin{threeparttable}
\resizebox{.48\textwidth}{!}{ 
\begin{tabular}{@{}c|c|c|c|c|c@{}}
\toprule
\multirow{2}{*}{Methods}                     & \multirow{2}{*}{Bitwidth}  & \multicolumn{2}{c|}{A5000} & \multicolumn{2}{c}{Orin Nano 8G}        \\ \cmidrule(l){3-6} 
                                            &                            & TPOT       & TTFT     & TPOT & TTFT  \\ \midrule \midrule
   -                                        & FP16                       & 22.73      & 197.80   & OOM  & OOM  \\ \midrule
 Quamba                                     & W8A8                       & 14.12      & 124.01   & OOM  & OOM  \\ \midrule
\multirow{3}{*}{\makecell{Quamba2\\(Ours)}} & W8A8                       & 12.61      & 122.33   & OOM  & OOM  \\
                                            & W4A8                       & 7.43       & 140.78   & 79.91  & 2088.03   \\ 
                                            & W4A16                      & 7.58       & 209.19   & 78.77  & 2316.23  \\ \midrule
\end{tabular}
}
\end{threeparttable}
\end{wraptable}

\vspace{-5pt}
\subsection{Latency and Model Size}
We test all methods on the A5000 for cloud applications and on the Orin Nano 8G for edge applications. Time-per-output-token (TPOT) and time-to-first-token (TTFT) are measured for a batch size of one, recorded in milliseconds (\emph{ms}). TTFT is profiled with 1024 input tokens. The results are shown in Table \ref{tab:quamba_latency} and Figure \ref{fig:mamba2_8b_memory}.
In the W8A8 setting, head-to-toe quantization of our framework improves the TPOT latency for Mamba2-8B by $1.80 \times$ (22.73 \emph{ms} \emph{vs.} 12.61 \emph{ms}), outperforming Quamba $1.61 \times$ \citep{chiang2024quamba} (22.73 \emph{ms} \emph{vs.} 14.12 \emph{ms}).
In the W4A8 configuration, Quamba2 achieves $3.89 \times$ less memory use, $1.39\times$ prefilling, and $3.05 \times$ faster generation speed for Mamba2-8B on A5000.
W4A8 and W4A16 slow down TTFT compared to W8A8 and FP16 due to dequantization overhead.
However, 4-bit weights bring latency benefits in the memory-bound generation stage.
Our approach allows the deployment of Mamba2-8B on Nano 8G with a speed of generating 13 tokens per second, while FP16 and W8A8 fail, as illustrated in Figure \ref{fig:mamba2_8b_memory} and Table \ref{tab:quamba_latency}.
For the SSD kernel, we load the 8-bit activations ($\bar{x}$, $\bar{A}$, $\bar{B}$, $\bar{C}$, $\bar{z}$) to reduce memory pressure and improve latency by $1.18 \times$, as shown in Table \ref{tab:ssd_latency}.
%
%

\subsection{Zero-shot Evaluation on Downstream Tasks}
We present the average accuracy for Quamba2 over five datasets: ARC-easy, ARC-challenge, PIQA, WinoGrande, and HellaSwag, allowing a fair comparison with MambaQuant \citep{xu2025mambaquant}.
The full evaluation is in the Appendix, where we follow the evaluation protocol in Mamba1 \citep{gu2023mamba}.
In contrast to Quamba \citep{chiang2024quamba}, when applied to Mamba1, our approach utilizes Hadamard transforms on input and output projections to increase quantization precision, thus enhancing accuracy for Mamba1.
As illustrated in Table \ref{tab:zero_shot}, our techniques \emph{sort-and-cluster} and \emph{per-state-group quantization} surpass clipping in Mamba2 \citep{dao2024transformers}. 
Our framework performs head-to-toe quantization, outperforming Quamba in latency and memory usage (refer to Table \ref{tab:quamba_latency} and \ref{tab:model_size}) for both W8A8 Mamba1 and Mamba2.
Quamba2 also outperforms MambaQuant in W4A8 Mamba1 and delivers real speedup on computing platforms.
Moreover, our framework supports W8A8, W4A8, and W4A16 precisions for both Mamba1 and Mamba2 with satisfactory accuracy and latency.

\begin{table}[h]
\centering
\small
\begin{minipage}{0.48\textwidth}
    \centering
    \caption{\textbf{(Zero-shot evaluation.)} We compare our framework with Quamba \citep{chiang2024quamba} and MambaQuant \citep{xu2025mambaquant} on the average accuracy of five zero-shot downstream tasks.}
    \resizebox{\textwidth}{!}{ 
    \begin{tabular}{@{}l|c|cc|cc@{}}
    \toprule
    \multirow{2}{*}{Bitwidth} & \multirow{2}{*}{Methods} & \multicolumn{2}{c|}{Mamba1}       & \multicolumn{2}{c}{Mamba2} \\
                              &                          & 1.4B            & 2.8B            & 2.7B            & 8B            \\ \midrule\midrule
    FP16                      & -                        & 58.6\%          & 62.2\%          & 62.4\%          & 70.8\%      \\ \midrule
    \multirow{3}{*}{W8A8}     & Quamba                   & 57.3\%          & 61.5\%          & 57.3\%          & 67.0\%      \\ \cmidrule{2-6} 
                              & MambaQuant               & \textbf{58.3}\% & \textbf{62.1}\% & -               & -           \\ \cmidrule{2-6} 
                              & Quamba2 (Ours)           & \underline{57.5}\%          & \underline{61.8}\%          & \textbf{62.1}\% & \textbf{69.9}\%      \\ \midrule
    \multirow{2}{*}{W4A8}     & MambaQuant               & \underline{54.3}\%          & \underline{58.5}\%          & -               & -           \\ \cmidrule{2-6} 
                              & Quamba2 (Ours)           & \textbf{56.7}\% & \textbf{61.0}\% & \textbf{61.4}\% & \textbf{69.4}\%      \\ \midrule
    W4A16                     & Quamba2 (Ours)           & \textbf{57.5}\% & \textbf{61.9}\% & \textbf{62.3}\% & \textbf{70.2}\%      \\ \bottomrule
    \end{tabular}
    }
    \label{tab:zero_shot}
\end{minipage}
\hfill
\begin{minipage}{0.48\textwidth}
    \centering
    \caption{\textbf{(Five-shot evaluation of Quamba2-8B on MMLU.)} We evaluate W4A8, W4A16, and W4A$X$-mixed on MMLU. Our W4A$X$-mixed model outperforms mixed by handcrafting (HC) and pure W4A8 models.}
    \renewcommand{\arraystretch}{1.25} 
    \resizebox{\textwidth}{!}{
    \begin{tabular}{@{}l|c|cc|c|c@{}}
    \toprule
    \multirow{2}{*}{Bitwidth} & \multirow{2}{*}{Method} & LAMB            & MMLU      & W4A\{$X$\} & \multirow{2}{*}{TTFT}   \\ 
                              &                         & (0-shot)        &  (5-shot) & (A8:A16)   &                         \\ \midrule\midrule
    FP16                      & -                       & 70.9\%          & 47.0\%    & -          & 197.80                  \\ \midrule
    W4A8                      & -                       & 68.8\%          & 41.2\%    & 56:0       & 140.78                  \\ 
    W4A16                     & -                       & 70.6\%          & 45.3\%    & 0:56       & 209.19                  \\ \midrule
    Mixed                     & HC-last                 & 68.3\%          & 42.1\%    & 42:14      & \multirow{3}{*}{158.36} \\
    Mixed                     & HC-first                & \underline{68.9}\%          & \underline{43.1}\%    & 42:14      &                         \\
    Mixed                     & \textbf{Auto}           & \textbf{69.1}\%          & \textbf{44.0}\%    & 42:14      &                         \\ \bottomrule
    \end{tabular}
    }
    \label{tab:mmlu-results}
\end{minipage}
\end{table}

\subsection{Evaluation on Large Multitasking Dataset}
\label{Sec: MMLU}
We evaluate W4A16 and W4A8 Quamba2-8B in the MMLU dataset \citep{hendrycks2020measuring}, a large multitasking dataset, covering 57 subject ranges at different difficulty levels.
Our study shows that previous quantization methods may overlook the generalizability of low bit-width models.
W4A8 strikes a balance between prefilling and generation speed but falls short in MMLU generalization, whereas W4A16 maintains a better generalization despite an increased prefilling latency, as shown in Table \ref{tab:mmlu-results}.
We handcraft two mixed-precision models that replace the last 14 layers and the first 14 layers with W4A16 denoted as HC-last and HC-first in the table, respectively.
However, they show marginal improvements on MMLU dataset.
To this end, we employ an evolutionary search approach to identify sensitive layers and assign W4A16 to these blocks.
The resulting mixed-precision model mitigates the loss of generalizability ($+2.9\%$) in the MMLU dataset, outperforming naive mixed-presion by handcrafting and pure W4A8 models, with only a 10\% increase in prefilling latency.

\vspace{-5pt}
\section{Ablation Studies}

\subsection{Ablation study on W4A8}
We conduct an ablation study on the W4A8 Quamba2-8B in Table \ref{tab:W4A8_ablation}.
In the W4A8 setting, it is essential to apply the Hadamard transform to the input of the output projection.
The model fails without applying Hadamard transforms.
However, due to the sensitivity of the SSM to quantization-induced errors, even with per-group quantization and GPTQ \citep{frantar2023gptq} (second-order information) applied on top of the Hadamard transform, the results remain unsatisfactory.
Our proposed methods \emph{per-state-group quantization} (PerSG) and \emph{sort-and-cluster} (SnC) address this issue in SSMs by quantizing the $x$, $B$, and $C$ in 8 bits with minimal accuracy drop.
It is noted that $x$ continues to be vulnerable to quantization errors in SSMs, consistent with the findings in \cite{chiang2024quamba}.
our sort-and-cluster technique outperforms clipping in addressing this issue (\emph{ref.} Table \ref{tab:zero_shot} and \ref{tab:zero_shot_full}).

\subsection{Ablation study on W4A16}

We study the impact of each component in the case of W4A16 Quamba2-8B (weight-only quantization, \emph{i.e.}, $\overline{W}X$), and show the results in Table \ref{tab:W4A16_ablation}.
The table demonstrates that the Hadamard transform combined with per-group weight quantization (PerG + Had.) yields greater accuracy than GPTQ \citep{frantar2023gptq} (PerG + GPTQ).
Our analysis indicates that the use of the Hadamard transform in the input of the out projection is crucial to narrowing the performance gap in weight-only quantization of SSMs.
Specifically, the Hadamard transform \emph{eliminates outliers} in the \emph{half-precision} activation, thereby avoiding the amplification of quantization errors from \emph{4-bit weights} by large outliers in the output projection such that $||W_{\text{out}}X_\text{out} - \overline{W}_{\text{out}}X^{\mathbf{H}}_\text{out} || < ||W_{\text{out}}X_\text{out} - \overline{W}_{\text{out}}X_\text{out} ||$.
By combining all methods (PerG + GPTQ + Had.), the W4A16 models close the performance gap between the half-precision on LAMBADA dataset.

\begin{table}[h]
\centering
\begin{minipage}{0.48\textwidth}
    \centering
    \caption{\textbf{(Ablation study on W4A8 Quamba2-8B.)} The accuracy on LAMBADA dataset is reported. (PerSG: per-state-group quantization for $B$ and $C$, SnC: sort-and-cluster for $x$, PerG: per-group weight quantization, GPTQ: \citet{frantar2023gptq}, and Had: Hadamard transforms)}
    \setlength{\tabcolsep}{2.5pt}
    \resizebox{\textwidth}{!}{
    \begin{tabular}{r|c|cc|ccc|c}
    \toprule
    \multirow{2}{*}{Size} & \multirow{2}{*}{Bitwidth}    & \multicolumn{2}{c|}{Weights} & \multirow{2}{*}{Had.}        & B/C        & x          & \multirow{2}{*}{Acc.} \\ 
                          &                              & PerG          & GPTQ        &        & PerSG      &  SnC     &                       \\ 
    \midrule\midrule
    \multirow{5}{*}[-2em]{8B}   & FP16                  & -             & -           & -          & -          & -          & 71.2\%                \\
                     \cmidrule{2-8}
                          & \multirow{5}{*}[-1em]{W4A8}  & \checkmark    &             &            &            &            & fail                  \\ \cmidrule{3-8}
                          &                              & \checkmark    &             & \checkmark &            &            & 53.8\%                \\ \cmidrule{3-8}
                          &                              & \checkmark    & \checkmark  & \checkmark &            &            & 55.1\%                \\ \cmidrule{3-8}
                          &                              & \checkmark    & \checkmark  & \checkmark & \checkmark &            & 60.7\%                \\ \cmidrule{3-8}
                          &                              & \checkmark    & \checkmark  & \checkmark & \checkmark & \checkmark & 68.8\%                \\ \bottomrule
    \end{tabular}
    }
    \label{tab:W4A8_ablation}
\end{minipage}
\hfill
\begin{minipage}{0.48\textwidth}
    \centering
    \caption{\textbf{(Ablation study on W4A16 Quamba2-8B.)} The accuracy on LAMBADA dataset is reported. The Hadamard transform eliminates  the large outliers from the half-precision activation, avoiding the amplification of quantization errors from 4-bit weights. (PerG: per-group quantization, GPTQ: \citet{frantar2023gptq}, and Had: Hadamard transforms)}
    \setlength{\tabcolsep}{7pt}
    \resizebox{.95\textwidth}{!}{
    \begin{tabular}{r|c|cc|c|c}
    \toprule
    \multirow{2}{*}{Size} & \multirow{2}{*}{Bitwidth}    & \multicolumn{2}{c|}{Weights} &  \multirow{2}{*}{Had.} & \multirow{2}{*}{Acc.} \\ 
                          &                              & PerG          & GPTQ         &                        &                       \\ 
    \midrule\midrule
    \multirow{5}{*}[-1em]{8B}   & FP16                  & -             & -            & -                       & 71.2\%                \\
                     \cmidrule{2-6}
                          & \multirow{4}{*}[-1em]{W4A16} & \checkmark    &             &                         & 64.7\%                \\ \cmidrule{3-6}
                          &                              & \checkmark    &             & \checkmark              & 69.6\%                \\ \cmidrule{3-6}
                          &                              & \checkmark    & \checkmark  &                         & 69.2\%                \\ \cmidrule{3-6}
                          &                              & \checkmark    & \checkmark  & \checkmark              & 71.2\%                \\ \bottomrule
    \end{tabular}
    }
    \label{tab:W4A16_ablation}
\end{minipage}
\end{table}

\begin{wraptable}{r}{0.5\textwidth} 
\centering
\caption{\textbf{(Ablation study on the embedding and output head.)} We experiment on quantizing the embedding and output head in addition to W4A8 blocks. The accuracy on the LAMBADA dataset is reported.}
\begin{tabular}{@{}r|c|cccc@{}}
\toprule
size                  & FP16 & \makecell{W4A8\\blocks} & \makecell{+ 4-bit\\lm\_head} & \makecell{+ 4-bit\\embed.} & + both \\ \midrule\midrule
130M & 43.7\%     & 37.6\%                   & 37.0\%                      & 33.4\%                    & 33.4\%    \\ \midrule
370M & 53.1\%     & 50.5\%                   & 50.3\%                      & 46.2\%                    & 46.6\%    \\ \midrule
2.7B & 69.5\%     & 65.8\%                  & 66.1\%                      & 66.0\%                     & 65.7\%    \\ \midrule
8B   & 70.9\%     & 68.5\%                  & 68.3\%                      & 69.0\%                     & 68.8\%    \\ \bottomrule
\end{tabular}
\label{tab:quantize_embed_head}
\end{wraptable}

\subsection{Quantizing the embedding and output head}

In Table \ref{tab:quantize_embed_head}, we perform an analysis of quantizing the embedding and output head in addition to W4A8 blocks.
As the weights in all layers are represented in 4-bit, the half-precision embedding layer and the output head become the memory bottlenecks, preventing the W4A8 models from being deployed to edge devices (\emph{ref.} Figure \ref{fig:mamba2_8b_memory} W4A8).
As a result, we experiment on quantizing the embedding and output head in addition to W4A8 blocks and show the results in Table \ref{tab:quantize_embed_head}.
We show that larger models present more resilience to quantizing both the embedding layer and the output head, as the accuracy on the LAMBADA dataset remains nearly unchanged.
This finding is particularly useful for deploying large models onto a device with limited memory.
Our framework provides different bit-width configurations (\emph{i.e.,} 4-bit and 8-bit) for the embedding layer and output head, addressing the needs for deploying large models on edge devices.
%


\section{Conclusion} \label{sec:conclusion}
We introduce Quamba2, a robust post-training quantization framework tailored for selective State Space Models, compatible with \textbf{W4A8}, \textbf{W4A16}, and \textbf{W8A8} on \textbf{Mamba1} and \textbf{Mamba2}. Using \emph{channel order preservation} and \emph{activation persistence} observed in SSMs, we propose \emph{sort-and-cluster} and \emph{per-state-group quantization} techniques for the quantization of 8-bit activation. Experiments demonstrate that Quamba2 surpasses previous methods, offering significant reductions in latency and memory for both cloud and edge applications, addressing deployment challenges for emerging SSM-based applications on various platforms.






\clearpage

\section*{Impact Statement}
This paper aims to enhance the efficiency of machine learning and expand the accessibility of large language models.
We find that the accuracy degradation is not negligible.
Despite this, the performance trade-off is acceptable given the significant improvements in latency and resource efficiency. 
Our work enables large language models to be deployed on resource-limited devices.
As a positive feature, our method may push the development of privacy-centric on-device applications, where sensitive data can be processed locally without relying on cloud services.
However, our work may also present challenges such as increased device resource usage and potential security vulnerabilities if the local devices are compromised.

\section*{Acknowledgments}
This work was supported in part by the ONR Minerva program, NSF CCF Grant No. 2107085, iMAGiNE - the Intelligent Machine Engineering Consortium at UT Austin, UT Cockrell School of Engineering Doctoral Fellowships, NSF CAREER Grant No. 2339084, and Taiwan's NSTC Grant No. 111-2221-E-A49-148-MY3.

\bibliographystyle{plainnat}   
\bibliography{biblio}          

\newpage

\appendix
\onecolumn

\section{Full Results for Six Zero-shot Downstream Tasks}
\label{Appendix:full_zero_shot_evaluation}

In Table \ref{tab:zero_shot_full}, we follow the evaluation protocol in Mamba \citep{gu2023mamba}, and report the accuracy for LAMBADA \citep{LambdataDataset}, WinoGrande \citep{WinoGrandeDataset}, PIQA \citep{PIQADataset} and ARC-easy \citep{arcDataset}, and the accuracy normalized by the sequence length for HellaSwag \citep{HellaSwag} and ARC-challenge \citep{arcDataset}. 
Given the slight variation in accuracy across runs, we present the average accuracy over five runs in each table.
Our frame work outperforms Quamba \citep{chiang2024quamba} in the Mamba1 backbone, providing with more quantization flavors such as W8A8, W4A8 and W4A16 for different use cases.
Our method also outperforms Quamba in the Mamba2 backbone, where we apply the clipping technique to Mamba2, by a large gap in the average accuracy.

\begin{table*}[h!]
\centering
\caption{\textbf{(Zero-shot accuracy.)} We evaluate our framework on six common sense tasks and report the average of five runs. Our framework surpass previous baseline, Quamba \citep{chiang2024quamba}, in average accuracy on both Mamba1 and Mamba2 backbones, with supporting more quantization flavors.}
\resizebox{48em}{!}{
\begin{tabular}{r|c|c|c|cccccc|c}
\toprule
Model & Size                  & Methods & Bitwidth                           & LA        & HS     & PIQA       & Arc-E   & Arc-C   & WG & Avg. \\
\midrule\midrule
\multirow{4}{*}[-5em]{Mamba}
& \multirow{2}{*}[-2em]{1.4B} 
                            &  -                                            &  FP16 & 64.9\% & 59.1\% & 74.2\% & 65.5\% & 32.8\% & 61.5\% & 59.7\% \\ \cmidrule{3-11}
&                           & Quamba                                        &  W8A8 & 61.4\% & 58.3\% & 72.7\% & 64.0\% & 32.3\% & 58.8\% & 57.9\% \\ 
                                        \cmidrule{3-11}
&                           & \multirow{3}{*}{\makecell{Quamba2\\(Ours)}}   & W8A8  & 62.3\% & 58.6\% & 73.1\% & 64.0\% & 32.2\% & 58.5\% & 58.1\% \\ 
&                           &                                               & W4A8  & 61.5\% & 57.6\% & 72.0\% & 63.0\% & 32.2\% & 58.7\% & 57.5\% \\ 
&                           &                                               & W4A16 & 63.6\% & 58.1\% & 72.6\% & 64.3\% & 32.4\% & 60.5\% & 58.5\% \\ 

    \cmidrule{2-11}
& \multirow{2}{*}[-2em]{2.8B} 
                            &  -                                            & FP16  & 69.1\% & 65.9\% & 75.6\% & 69.2\% & 35.8\% & 63.0\% & 63.1\% \\ \cmidrule{3-11}
&                           & Quamba                                        & W8A8  & 65.4\% & 65.1\% & 74.2\% & 68.9\% & 35.9\% & 62.6\% & 62.0\% \\ 
                                        \cmidrule{3-11}
&                           & \multirow{3}{*}{\makecell{Quamba2\\(Ours)}}   & W8A8  & 65.7\% & 65.4\% & 74.5\% & 68.9\% & 36.7\% & 61.8\% & 62.2\% \\ 
&                           &                                               & W4A8  & 63.5\% & 64.9\% & 74.2\% & 68.2\% & 35.3\% & 62.2\% & 61.4\% \\ 
&                           &                                               & W4A16 & 66.0\% & 65.3\% & 74.6\% & 69.2\% & 36.6\% & 63.6\% & 62.6\% \\ 
\midrule
\multirow{9}{*}[-6em]{Mamba2}
& \multirow{3}{*}[-2em]{1.3B} 
                            &  -                                            & FP16  & 65.6\% & 59.9\% & 73.3\% & 64.1\% & 33.3\% & 60.8\% & 59.5\% \\ \cmidrule{3-11}
&                           & Quamba                                        & W8A8  & 49.8\% & 58.5\% & 71.2\% & 61.9\% & 32.1\% & 58.1\% & 55.2\% \\ 
                                        \cmidrule{3-11}
&                           & \multirow{3}{*}{\makecell{Quamba2\\(Ours)}}   & W8A8  & 62.0\% & 59.2\% & 72.5\% & 63.4\% & 32.7\% & 60.0\% & 58.3\% \\ 
&                           &                                               & W4A8  & 61.0\% & 58.8\% & 72.4\% & 62.7\% & 32.6\% & 59.1\% & 57.7\% \\ 
&                           &                                               & W4A16 & 64.3\% & 59.2\% & 72.6\% & 63.8\% & 33.1\% & 60.3\% & 58.9\% \\ 
    \cmidrule{2-11}
& \multirow{3}{*}[-2em]{2.7B} 
                            &  -                                            & FP16  & 69.5\% & 66.6\% & 76.4\% & 69.5\% & 36.4\% & 64.2\% & 63.8\% \\ \cmidrule{3-11}
&                           & Quamba                                        & W8A8  & 52.4\% & 60.4\% & 71.6\% & 62.9\% & 33.7\% & 58.0\% & 56.5\% \\ 
                                        \cmidrule{3-11}
&                           & \multirow{3}{*}{\makecell{Quamba2\\(Ours)}}   & W8A8  & 66.1\% & 65.5\% & 74.4\% & 68.4\% & 37.1\% & 63.7\% & 62.5\% \\ 
&                           &                                               & W4A8  & 65.6\% & 65.1\% & 74.7\% & 68.1\% & 36.1\% & 62.8\% & 62.1\% \\ 
&                           &                                               & W4A16 & 68.8\% & 65.6\% & 75.5\% & 68.6\% & 36.6\% & 64.9\% & 63.3\% \\ 
    \cmidrule{2-11}
& \multirow{3}{*}[-2em]{8B} 
                            &  -                                            & FP16  & 70.9\% & 77.7\% & 79.7\% & 76.0\% & 48.0\% & 72.0\% & 70.7\% \\ \cmidrule{3-11}
&                           & Quamba                                        & W8A8  & 54.0\% & 74.6\% & 77.1\% & 73.5\% & 44.2\% & 65.5\% & 64.8\% \\ 
                                        \cmidrule{3-11}
&                           & \multirow{3}{*}{\makecell{Quamba2\\(Ours)}}   & W8A8  & 69.8\% & 77.8\% & 79.1\% & 75.9\% & 46.9\% & 69.0\% & 69.8\% \\ 
&                           &                                               & W4A8  & 68.8\% & 77.1\% & 79.1\% & 75.0\% & 46.0\% & 68.7\% & 69.1\% \\ 
&                           &                                               & W4A16 & 71.2\% & 76.8\% & 79.1\% & 75.2\% & 45.9\% & 70.8\% & 69.8\% \\ 
\bottomrule

\end{tabular}
}
\label{tab:zero_shot_full}
\end{table*}

\clearpage

\section{Evaluation Results on Generation Tasks}

\begin{wraptable}{r}{0.33\textwidth} 
\vspace{-20pt}
\small
\centering
\caption{\textbf{(Generation tasks.)} We evaluate Mamba2-8B with different precisions on the generation tasks.}
\label{tab:quamba2_generation}
\resizebox{.3\textwidth}{!}{ 
\begin{tabular}{l|c|c}
\toprule
Bit-width  & {NQ} & {SquadV2} \\
\midrule
FP16   & 17.2 & 51.9 \\
W8A8   & 15.0 & 43.6 \\
W4A8   & 14.2 & 45.9 \\
W4A16  & 16.6 & 50.7 \\
W4AX   & 14.9 & 47.4 \\
\bottomrule
\hline
\end{tabular}
}
\end{wraptable}

We evaluate Mamba2-8B with all bit-widths on the generation-based tasks Natural Questions (NQ) (exact match) \citep{lee2019latent} and SquadV2 (F1) \citep{rajpurkar2018know} on the open-source LM-EVAL \citep{eval-harness}.
We show the results in Table \ref{tab:quamba2_generation}.
The W4A16 model closely matches the FP16 model, whereas the W4A8 and W8A8 models, with 8-bit SSM states, preserve the meaningful generation outputs.
We show that the searched W4A$X$ also improves the generation scores and outperforms the W4A8 model.
This result reveals an interesting observation that cached SSM states are \emph{redundant}, which can be \emph{carefully} quantized to 8 bits.
Our framework supports 8-bit SSM states for W4A8 and W8A8 models and improves their generation speeds with large batch-size inputs, as the cached states are the major memory and latency bottlenecks.
Please refer to Section \ref{sec:large_batch_size} for more details.

\section{Implementation and Evaluation Details of Quamba2 Framework}
\label{sec: Details of Quamba2 Framework}

\paragraph{Quantization setup.}
The calibration set is constructed by randomly sampling 512 sentences from the Pile dataset \citep{PileDataset}, where we fixed the random seed in the sampling process.
We collect the \emph{static} scaling factors for each operator based on the absolute maximum value observed from the calibration set to quantize the activations and cached SSM states in both W4A8 and W8A8 settings.
The same scaling factors are applied in all our experiments.

\paragraph{Implementation.}
We implement our framework based on \texttt{CUTLASS} \citep{Thakkar_CUTLASS_2023}, \texttt{vLLM} \citep{kwon2023efficient}.
Our 4-bit and 8-bit matrix multiplication (matmul) kernels are adapted from \citep{xiao2023smoothquant, frantar2024marlin, zhang2024qqq, cuda_hgemv, cuda_hgemm}.
We implement W4A8 and W4A16 fused matmul-transpose kernels for the Mamba1 architecture.
We apply GPTQ \citep{frantar2023gptq} to the projection layers in the 4-bit weight settings.
Quantization is integrated and adapted to the CUDA kernels of both the fast Hadamard transform \citep{tridao_had} and causal convolution \citep{tridao_conv}.
Furthermore, the selective scan and SSD kernels \citep{gu2023mamba, dao2024transformers} are modified to accommodate inputs with quantized weights, activations, and their scaling factors.

\paragraph{Latency and model size profiling.}
We evaluate all methods on the A5000, a widely used GPU for AI workloads with 24GB of memory, emulating the setting for cloud applications.
For edge applications, we profile all methods on the Nvidia Orin Nano 8G.
We perform a few warm-up iterations and then report the average latency of the next 100 iterations.
We report the size of the model that includes all quantized parameters and buffers for calculation.

\section{Details for Mixed Precision Quamba2} 
\label{Appendix: Precsion Config}
\begin{figure}[h!]
    \centering
    \includegraphics[width=\textwidth]{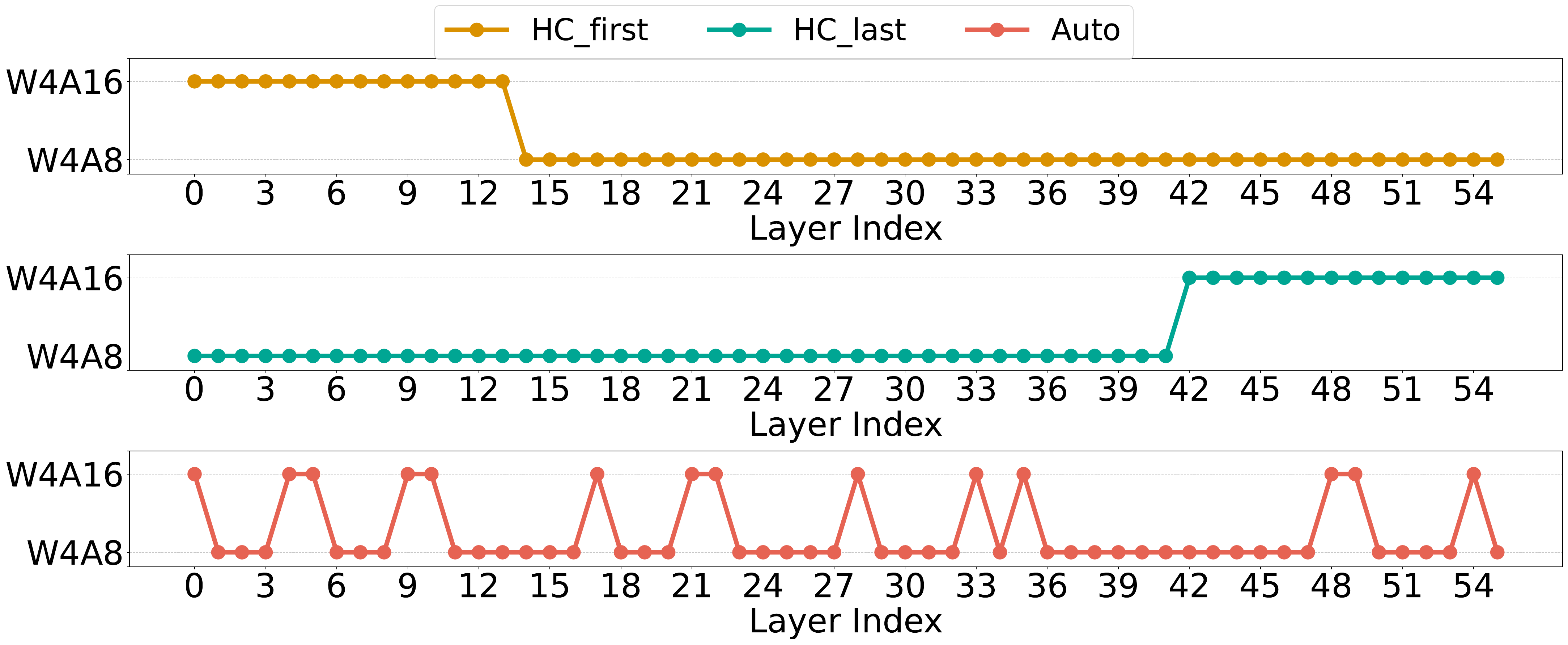} 
    \caption{\textbf{(The layer-wise bit-width for Quamba2-8B-W4A$X$.)} We search the bit-width for Quamba2-8B-W4A$X$ (the last row in red), which outperforms the handcraft counterparts shown in the first (HC\_first) and the second (HC\_last) rows.}
    \label{fig:prec_search}
\end{figure}

In Table \ref{tab:mmlu-results}, we outline the generalizability issue when utilizing the precision of W4A8 only. We show that our W4A$X$ mixed-precision models mitigate accuracy degradation while incurring only a marginal latency overhead. Figure \ref{fig:prec_search} visualizes the detailed layer-wise bit-width configuration of Quamba2-8B-W4A$X$.

\paragraph{The handcrafted mixed-precision models.}
We explored two types of handcrafted (HC) configurations, referred to as HC\_first and HC\_last, where we apply W4A16 blocks at the beginning and end of the network, respectively. 
Handcrafted configurations only deliver marginal improvements in the average accuracy (approximately 1\% on MMLU), and still fall behind in the upper bound scenario, where all blocks utilize the precision of W4A16, as shown in Table \ref{tab:mmlu-results}.

\paragraph{The automated W4A$X$ models.} 
We implement evolutionary search to identify the best mix of precision levels \citep{guo2020single}. We set the population size to 40 and the number of generations to 5. In each generation, the top performing half of the candidates are retained, with 10 mutation and crossover operations applied, respectively, to generate new candidate precision configurations. 
The search algorithm identifies the sensitive blocks and assigns W4A16 to these blocks.
This automated approach searches the best mix-precision configurations and balances between the precision and performance.
Our W4A$X$ models addresses the performance gaps in the MMLU dataset, as shown in Table \ref{tab:mmlu-results}, compared to naive mixed-precision and pure W4A8 models.

\vspace*{5pt}

\begin{wrapfigure}{r}{0.5\textwidth}
    \vspace{-15pt}
    \centering
    \includegraphics[width=.4\textwidth]{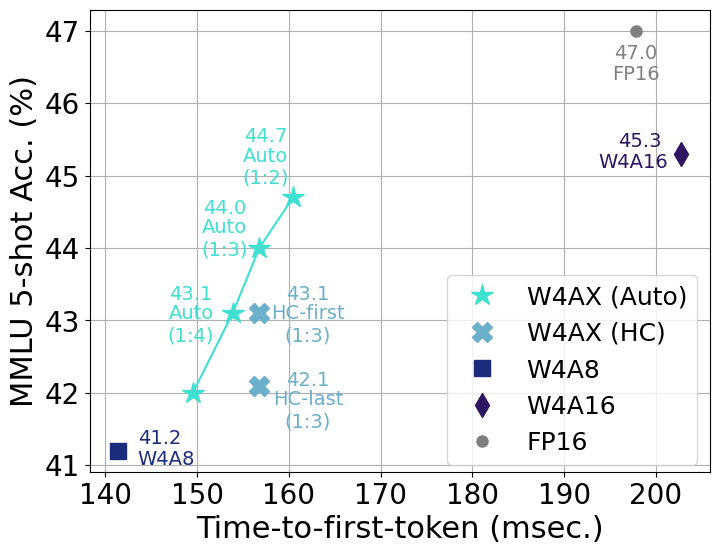}
    \caption{(\textbf{Pareto front analysis for mixed-precision models.}) Our W4A$X$ models by searched (Auto) outperform naive handcrafted (HC) models in MMLU accuracy and prefilling latency trade-off.}
    \label{fig:bitwidth_tradeoff}
\end{wrapfigure}

\paragraph{Analysis on W4A$X$ latency and accuracy trade-off.}
In Figure \ref{fig:bitwidth_tradeoff}, we show our W4A$X$ models outperform naive handcrafted models in MMLU \citep{hendrycks2020measuring} five-shot accuracy, and place at the Pareto-frontier of prefilling latency (time-to-first-token, TTFT) trade-off.
In this experiment, we adjusted the ratios of W4A16 and W4A8 (\emph{e.g.,} 1:2) in Quamba2-8B and used evolutionary search to find the mixed precision configuration.
As shown in the figure, the searched W4A$X$ models in different ratios improve the accuracy of the 5-shot evaluation on MMLU compared to W4A8, introducing marginal pre-filling latency overheads (\emph{i.e.}, $140.7$ \emph{vs.} $158.3$ ms).
Moreover, the automatic designed W4A$X$ models by our search algorithm are above naive handcrafted W4A$X$ models in accuracy.
This finding highlights the \emph{challenges} of designing \emph{mixed-precision} models for SSMs, as well as the limits of \emph{generalization} \citep{zhao2024qspec, kumar2025scaling} of low-bit SSMs on large-scale datasets.
We expect more advanced search algorithms to address the generalization issue in the future.

\vspace*{20pt}
\section{Investigating Memory and Latency with Large Batch Sizes}
\label{sec:large_batch_size}

\begin{figure*}[h]
    \centering
    \includegraphics[width=\textwidth]{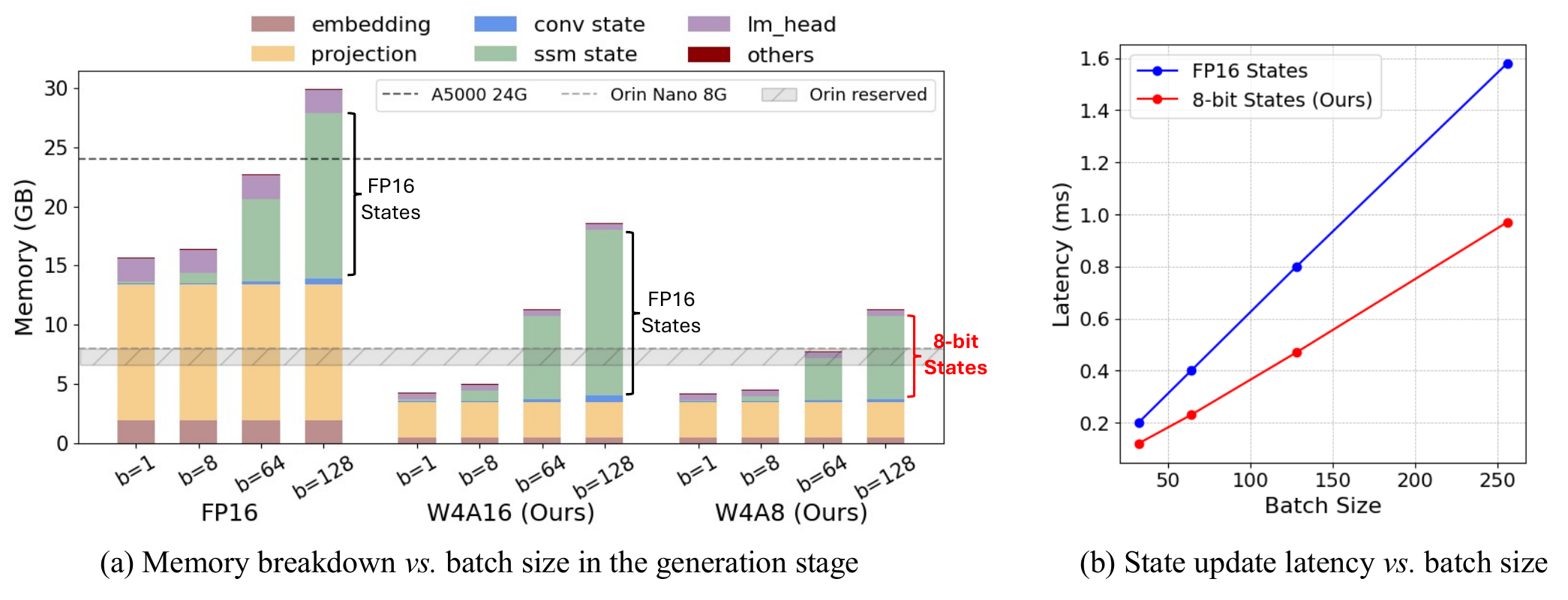}
    \caption{(\textbf{Large batch inputs.}) The cached states grow linearly with respect to the input batch size. For a batch size of 128, half-precision cached states use most of the memory (a), making state loading and updating the bottleneck during generation. Our framework (W4A8) compresses the states to 8-bit, thereby reducing the total memory and generation latency (b) with large batch size inputs for cloud-based applications.}
    \label{fig:ssm_large_batch}
\end{figure*}

\vspace{5pt}
\paragraph{The cached state sizes.}
Although the constant state nature of SSMs, the cached states grow \emph{linearly} with respect to the input \emph{batch size}. We show theoretical memory breakdowns versus batch size in Figure \ref{fig:ssm_large_batch} (a). As the batch size increased, cached states occupied most of the total memory, making state loading and updating the bottleneck during generation. Our framework (W4A8) compresses and updates the states with 8-bit, thus decreasing overall memory usage and generation latency for cloud applications with large batch sizes.

\vspace*{15pt}
\paragraph{Quantizing cached SSM states.}
We reduce generation latency by quantizing the cached SSM states to 8-bit for W4A8 and W8A8 models.
Since the cached SSM states follow the head reordering and channel grouping indices from the SSM input $x$ (\emph{ref.} Figure~\ref{fig:clustering}), we apply the same $m$ head and $n$ channel groups to quantize each SSM state before caching them in memory.
This finding eliminates the need for additional online reordering of SSM states and only requires calibrating the SSM quantization scales.
Our approach introduces $\texttt{dstate} \times m \times n$ floating-point scales with minimal latency overhead, while significantly reducing the state update latency, as shown in Figure~\ref{fig:ssm_large_batch} (b).

\vspace*{5pt}
\begin{figure*}[h!]
    \centering
    \includegraphics[width=\textwidth]{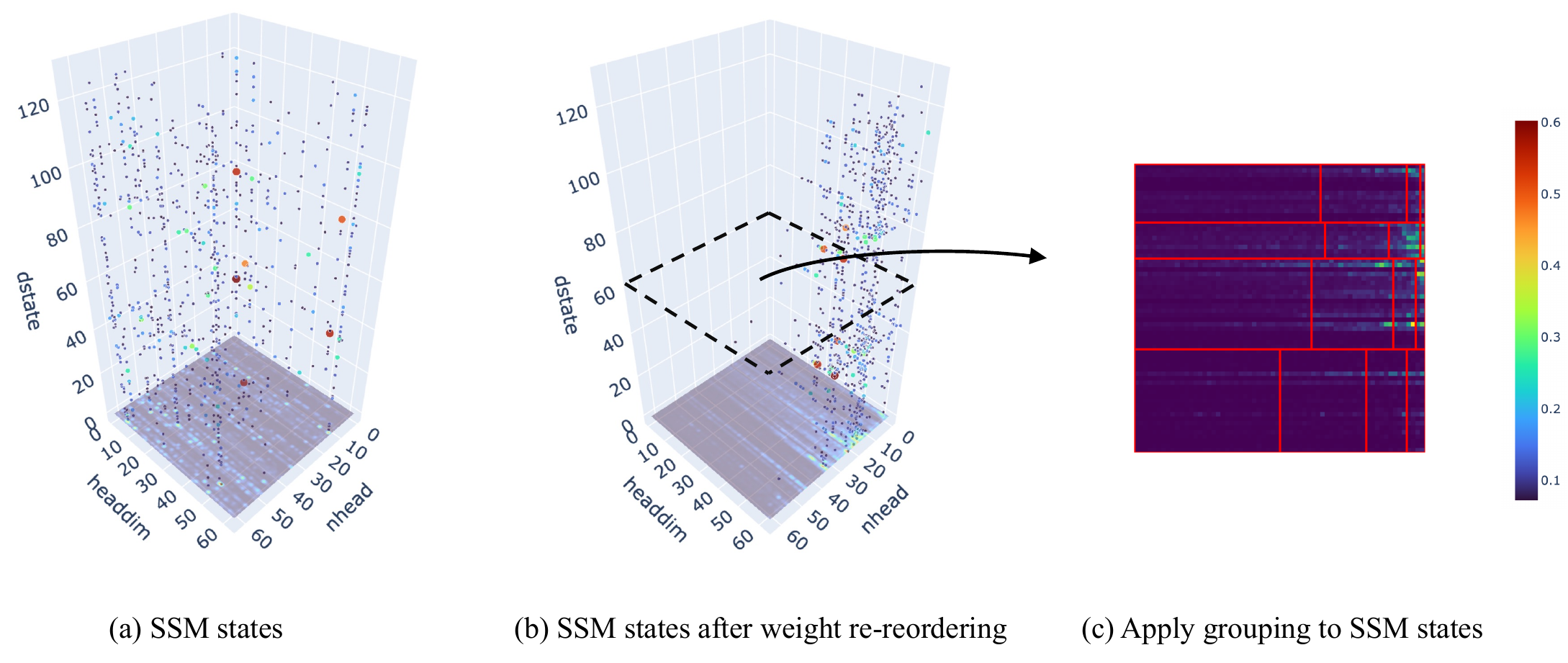}
    \caption{(\textbf{SSM states.}) The states are quantized before cached in memory. We apply the same $m$ head and $n$ channel groups from the SSM input $x$ to SSM states (b-c).}
    \label{fig:ssm_state_quantization}
\end{figure*}

\begin{figure}[h]
\centering
\begin{minipage}{0.48\textwidth}
    \centering
    \includegraphics[width=0.85\linewidth]{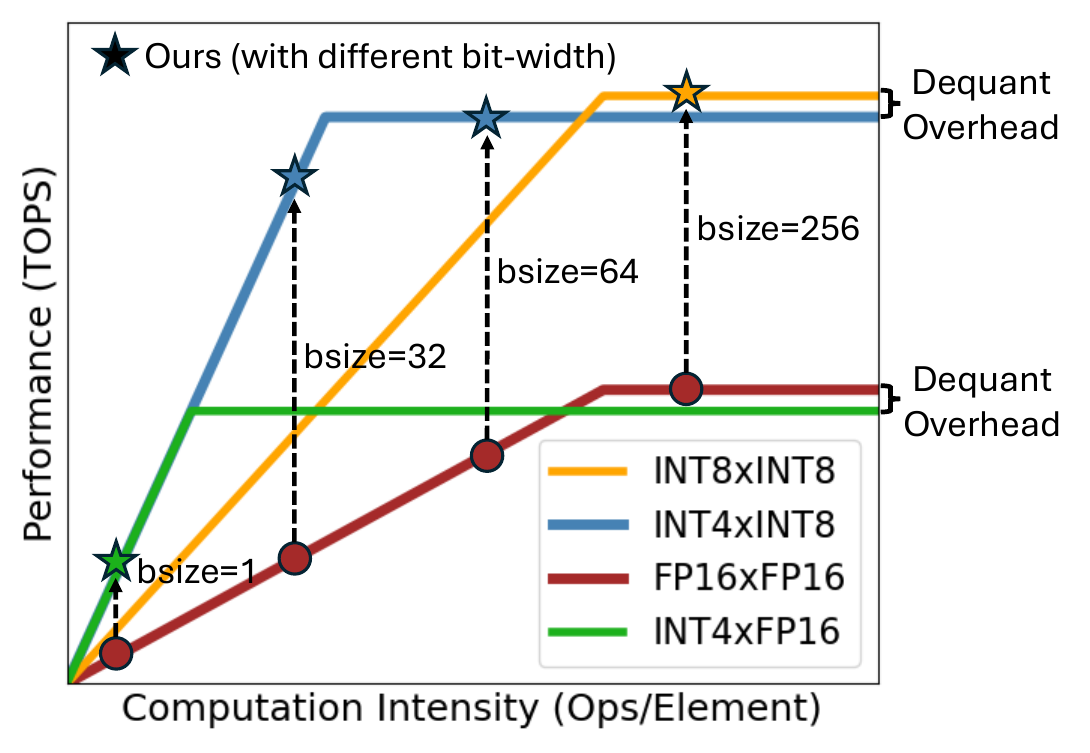}
    \caption{\textbf{(Roofline model of A5000.)}}
    \label{fig:roofline}
\end{minipage}
\hfill
\begin{minipage}{0.48\textwidth} 
    \centering
    \caption{\textbf{(Mamba2-8B TPOT on A5000 24GB.)} We compress the cached SSM states with 8-bit, enabling larger batch size inputs under the same memory constraints. We report latency in milliseconds (\emph{ms}). OOM denotes out-of-memory.}
    \label{tab:ssm_large_batch_latency}
    \small 
    \begin{tabular}{cccccc}
    \toprule
    \textbf{Bitwidth} & \textbf{b=1} & \textbf{b=32} & \textbf{b=64} & \textbf{b=128} & \textbf{b=256} \\
    \midrule\midrule
    \textbf{FP16}  & 22.73 & 35.74 & 49.63 & OOM & OOM \\
    \textbf{W8A8}  & 12.61 & 23.83 & 30.82 & 44.85 & 79.65 \\
    \textbf{W4A8}  & 7.43  & 15.05 & 24.65 & 44.54 & 85.26 \\
    \textbf{W4A16} & 7.58  & 20.58 & 38.48 & 74.25 & OOM \\
    \bottomrule
    \end{tabular}
\end{minipage}

\end{figure}

\vspace{10pt}
\paragraph{The roofline model.}
We show the roofline model of A5000 GPU in Figure \ref{fig:roofline} (\texttt{w-bit}$\times$\texttt{a-bit} in the figure), and profile the generation latency (\emph{i.e.,} time-per-output-token, TPOT) of Mamba2-8B on a A5000 with different batch sizes in Table \ref{tab:ssm_large_batch_latency}.
When the input batch size is small (\emph{e.g.,} \texttt{b=1} in the table), the generation is memory-bound and therefore loading 4-bit weights (\emph{e.g.,} W4A8 and W4A16) improves the roofline model.
As the batch size increased (\emph{e.g.,} \texttt{b=64} in the table), the W4A16 models are bounded by hardware performance in terms of trillions of operations per second (TOPS).
In contrast, the W4A8 and W8A8 models leverage 8-bit computation and deliver better TOPS.
The ultimate TOPS of W4A8 is lower than W8A8 due to the extra steps for dequantizing weights from 4-bit to 8-bit (\emph{e.g.,} \texttt{b=256} in the table).
Our framework supports \textbf{W8A8}, \textbf{W4A8}, and \textbf{W4A16} that are at the frontier of the roofline model to satisfy the deployment needs of most applications for both Mamba1 and Mamba2. 

\vspace{5pt}

\begin{wrapfigure}{r}{0.45\textwidth}
    \vspace{-0pt}
    \centering
    \includegraphics[width=.4\textwidth]{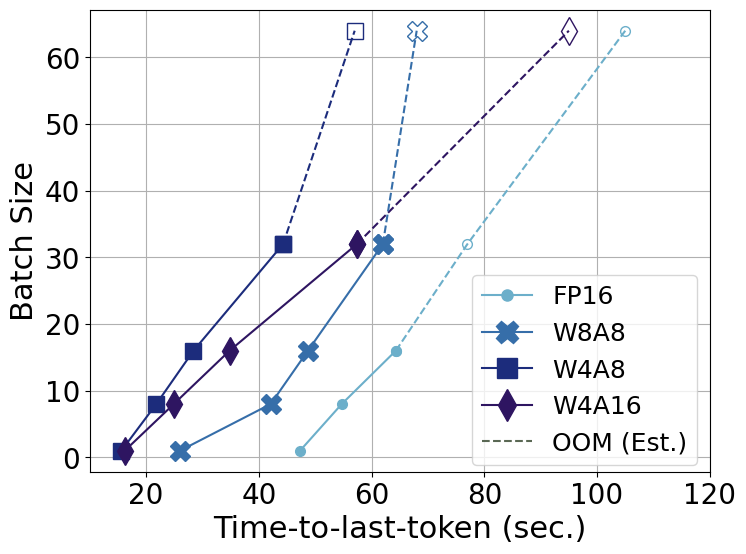}
    \caption{(\textbf{Batch size \emph{vs.} time-to-last-token.}) W4A8 is suited for most applications serving with general batch sizes among all supported bit-widths.}
    \label{fig:pareto_batch_ttlt}
\end{wrapfigure}

\paragraph{Batch size \emph{vs.} time-to-last-token latency across bit-widths.}
Figure \ref{fig:pareto_batch_ttlt} shows the time-to-last-token (TTLT) of Mamba2-8B quantized with different bit-widths (\emph{e.g.,} W8A8, W4A8, and W4A16) supported by our framework on a A5000.
We vary the batch size of the input from 1 to 64, and profile the end-to-end latency of pre-filling 2024 tokens and generating 2048 tokens (\emph{i.e.,} TTLT).
The latency is estimated for the batch sizes that empirically do \emph{not} fit A5000 and is represented with \emph{dashed} lines with \emph{unfilled} markers.
We show that the W4A8 Mamba-8B model is suited for most latency-sensitive applications, serving with general batch sizes (\emph{i.e.,} range from 1 to 64) on both cloud and edge devices.
In contrast, W4A16 serves as a better option for personal applications (\emph{i.e.,} batch size equal to one) on mobile platforms as it features higher average accuracy (\emph{ref.} Table \ref{tab:zero_shot_full} and \ref{tab:mmlu-results}).
For large batch size (\emph{i.e.,} greater than 128), the W8A8 model delivers the highest performance in terms of latency. 
Our framework supports all options on the frontier of the roofline model, as shown in Figure \ref{fig:roofline}.

\section{Accuracy-latency Trade-off}
\label{sec:Pareto front analysis}

\begin{figure*}[h]
    \centering
    \includegraphics[width=\textwidth]{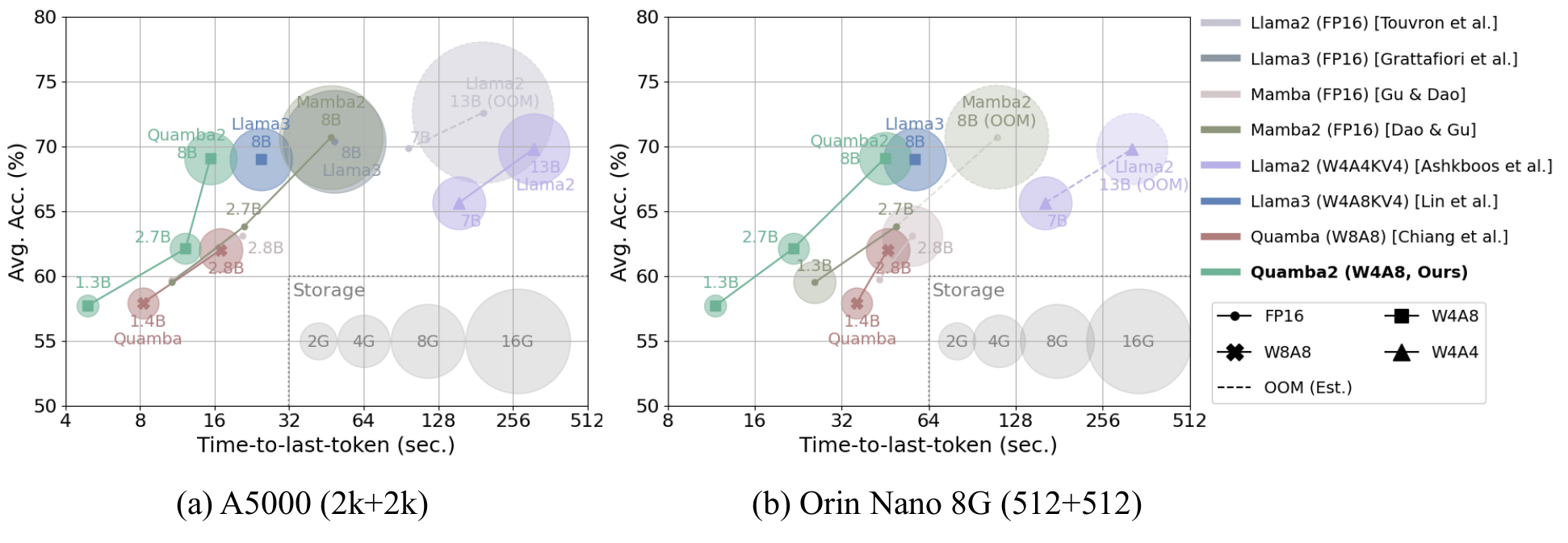}
    \caption{(\textbf{Pareto front analysis for accuracy \emph{vs.} latency.)} Quamba2 models (green) are on the Pareto front over other low bit-width SSM (red) and Transformer (purple) baselines, while also featuring lower memory footprints as evidenced in the size of the circle.}
    \label{fig:a5000_nano_ttlt}
\end{figure*}

\paragraph{Accuracy \emph{vs.} latency across backbone models.}
Figure \ref{fig:a5000_nano_ttlt} illustrates the average accuracy across six zero-shot tasks (y-axis) versus latency (x-axis, in log-scale) on a cloud-based A5000 GPU (a) and an Orin Nano 8G (b). 
We profile TTLT (time-to-last-token) in seconds (\emph{sec.}), with 2K input tokens and 2K generated tokens on the A5000 GPU.
For the Orin Nano 8G, we profile the TTLT with prefilling of 512 input tokens and 512 generated tokens.
For QuaRot \citep{ashkboos2024quarot}, we use the official implementation and profile latency for Llama2 \citep{touvron2023llama}.
We profile Llama3 \citep{grattafiori2024llama} and use the official QServe implementation \citep{lin2024qserve} to quantize it to W4A8KV4.
We note that the latencies and memory denoted with \emph{dashed} lines and circles are merely estimated.
For example, FP16 Llama2 13B is too large for the A5000's 24GB GPU memory, and W4A4 Llama2 13B also exceeds the capacity of Orina Nano.
Quamba2 models are on the Pareto frontier and offer the best trade-off between average accuracy and latency, as well as smallest memory footprints, outperforming other low bit-width SSM and Transformer baselines.

\begin{figure}[h!]
\centering
\includegraphics[width=.75\linewidth]{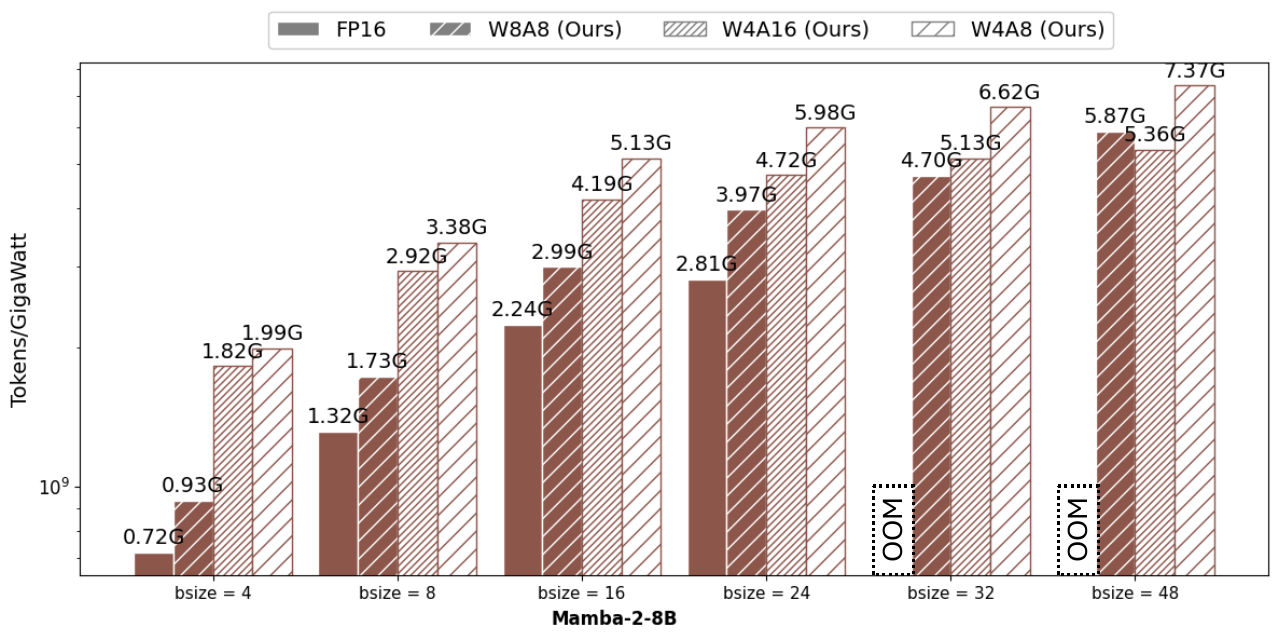}
\caption{
\textbf{(Energy efficiency analysis on A5000.)}
Each request is prefilled with 1024 tokens and generates 1024 new tokens.
The energy efficiency is measured with token per Gigawatt.
}
\label{fig:token_gwatts}
\end{figure}

\section{Energy Profiling}
\label{Appendix: Energy Profiling}

\begin{wraptable}{r}{0.48\textwidth}  
\vspace{-38pt}
\centering
\caption{\textbf{(Energy profiling on Nano.)} Joules per request (Js/req.) is reported. Each request is prefilled with 512 tokens and 512 generated tokens. Lower is better ($\downarrow$). }
\setlength{\tabcolsep}{10pt}
\begin{tabular}{@{}r|c|c@{}}
\toprule
{Method} & 
{Bit-width} & 
{Mamba2-8B} \\
\toprule
-          & FP16                                       & OOM     \\ 
\midrule
\multirow{2}{*}{\makecell[r]{\textbf{Quamba2} \\ \textbf{(Ours)}}} 
               & W4A8       & 231.23 \\
               & W4A16      & 225.46 \\
\bottomrule
\end{tabular}
\label{tab:energy_on_nano}
\end{wraptable}


To assess the practical efficiency of Quamba2, we conduct energy profiling on an A5000 GPU and an Orin Nano 8G, both of which are representative of cloud and edge platforms. 
On Orin Nano, each request is prefilled with 512 tokens and generates 512 new tokens, where we record the total energy consumption in joules per request (Js/req.).
As shown in Table~\ref{tab:energy_on_nano}, full-precision (FP16) models exceed the device's 8~GB memory limit, resulting in out-of-memory (OOM) errors during inference. 
In contrast, quantized models like W4A8 and W4A16 handle each request with 231.23~J and 225.46~J, respectively, while remaining deployable.

On cloud GPUs, we evaluate energy efficiency in terms of \textit{tokens per Gigawatt} to align with the industrial computing power metric.
In Figure~\ref{fig:token_gwatts}, we visualize the energy efficiency of Mamba2-8B on an A5000 GPU with 24GB memory under different batch sizes and quantization settings.
Each request is prefilled with 1024 tokens and generates 1024 new tokens.
Quamba2 consistently achieves higher throughput-per-energy than the FP16 model. 
This shows Quamba2 as an effective deployment framework for both resource-constrained and cloud-serving scenarios.


\onecolumn

\end{document}